\pdfoutput=1
\documentclass[11pt]{article}

\usepackage[final]{coling}

\usepackage{times}
\usepackage{latexsym}

\usepackage[T1]{fontenc}
\usepackage[utf8]{inputenc}

\usepackage{microtype}

\usepackage{inconsolata}

\usepackage{graphicx}

\usepackage{url}
\usepackage{booktabs}
\usepackage{multirow}
\usepackage{amsfonts}
\usepackage{amssymb}
\usepackage{bbding}
\usepackage{dashrule}
\usepackage{subfigure}
\usepackage{colortbl}
\usepackage{amsmath}

\title{ToolEyes: Fine-Grained Evaluation for Tool Learning Capabilities\\ of Large Language Models in Real-world Scenarios}

\author{
    \bf{\normalsize
    Junjie Ye$^{1}$, \ \ Guanyu Li$^{1}$, \ \ Songyang Gao$^{1}$, \ \ Caishuang Huang$^{1}$, \ \ Yilong Wu$^{1}$,}
    \bf{\normalsize \ \ Sixian Li$^{1}$,\ \ }\\
    \bf{\normalsize Xiaoran Fan$^{1}$, \ \ Shihan Dou$^{1}$, \ \ Tao Ji$^{1}$, \ \ }
    \bf{\normalsize Qi Zhang$^{1,3,4}$\thanks{Corresponding authors.}, \ \ Tao Gui$^{2,5*}$, \ \ Xuanjing Huang$^{1,3}$} \\ 
  {$^1$ \normalsize School of Computer Science, Fudan University} \\
  {$^2$ \normalsize Institute of Modern Languages and Linguistics, Fudan University} \\
  {$^3$ \normalsize Key Laboratory of Intelligent Information Processing, Fudan University, Shanghai, China} \\
  {$^4$ \normalsize Research Institute of Intelligent Complex Systems, Fudan University}\\
  {$^5$ \normalsize Pengcheng Laboratory}\\
  \texttt{\normalsize jjye23@m.fudan.edu.cn, \{qz, tgui\}@fudan.edu.cn} \\
  }

\begin{document}
\maketitle

\begin{abstract}
Existing evaluations of tool learning primarily focus on validating the alignment of selected tools (e.g., various APIs) for large language models (LLMs) with expected outcomes. However, these approaches rely on a limited set of scenarios where answers can be pre-determined.
Furthermore, a \emph{sole} emphasis on outcomes disregards the complex capabilities required for LLMs to effectively use tools.
To tackle this issue, we propose \emph{ToolEyes}, a fine-grained system tailored for the evaluation of the LLMs' tool learning capabilities in authentic scenarios. The system meticulously examines seven real-world scenarios, analyzing five dimensions crucial to LLMs in tool learning: \emph{format alignment}, \emph{intent comprehension}, \emph{behavior planning}, \emph{tool selection}, and \emph{answer organization}.
Additionally, ToolEyes incorporates a tool library boasting approximately 600 tools, serving as an intermediary between LLMs and the physical world. Evaluations involving ten LLMs across three categories reveal a preference for specific scenarios and limited cognitive abilities in tool learning.
Intriguingly, expanding the model size even exacerbates the hindrance to tool learning.
The code and data are available at~\url{https://github.com/Junjie-Ye/ToolEyes}.
\end{abstract}

\section{Introduction}
\label{sec:intro}

\begin{figure}[!t]
    \centering
    \vspace{5mm}
    \includegraphics[width=0.9\linewidth]{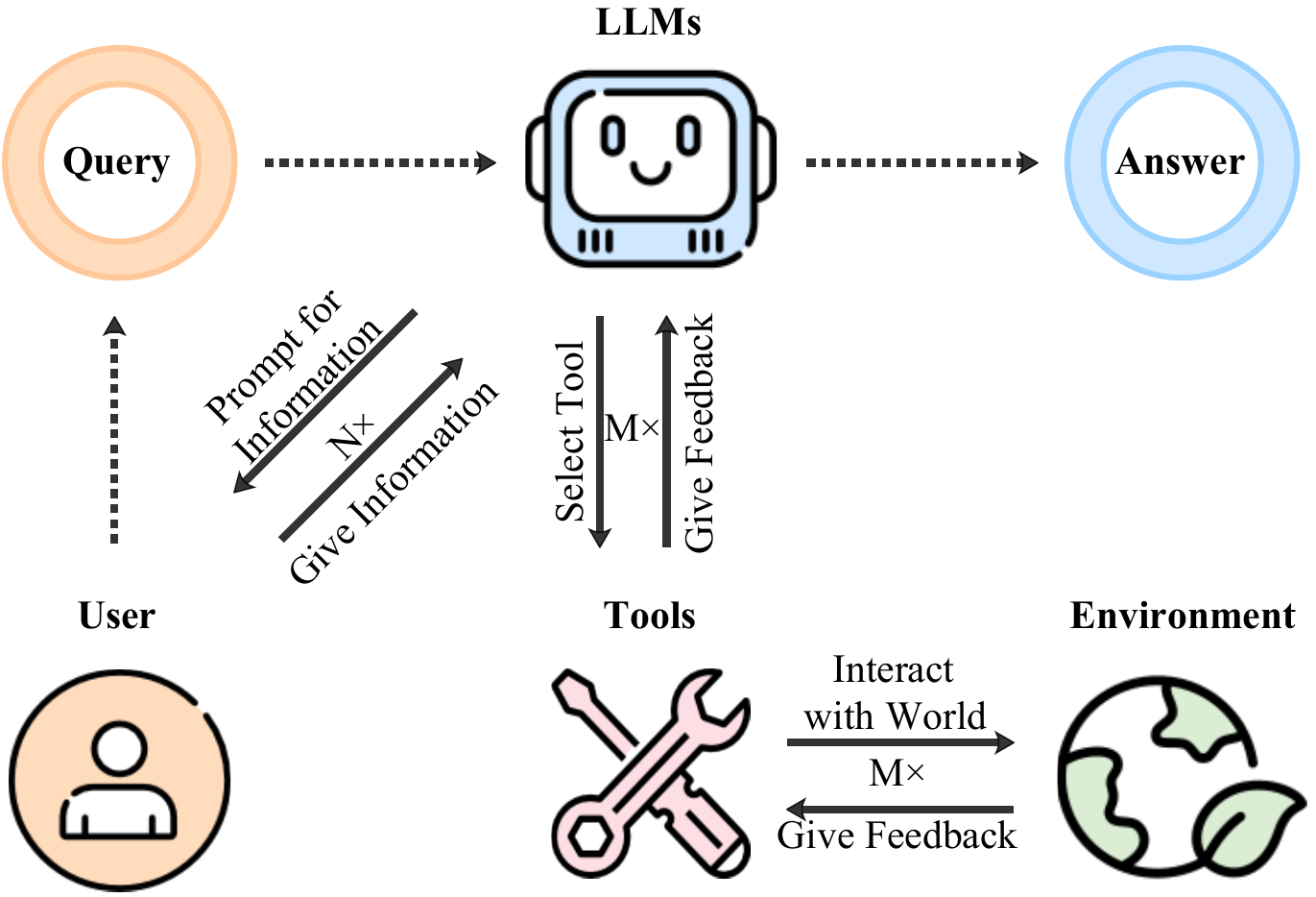}
    \caption{Illustration of tool learning. To address user queries, LLMs must analyze user requirements, utilize appropriate tools, and extrapolate feedback from the environment. Each stage in this process plays a crucial role in shaping the formulation of the answer.
    }
    % \vspace{-4mm}
    \label{fig:tool-learning}
\end{figure}

Large language models (LLMs)~\cite{GPT-3, Claude, LLaMA} represent a significant opportunity for advancing artificial intelligence (AI) owing to their remarkable performance across a diverse set of general-purpose tasks~\cite{analy-ye, analysis-chen, eval-survey}. To further bolster the model's capacity to meet real-world demands, researchers are actively exploring tool learning through the integration of external tools (e.g., various APIs)~\cite{foundation, augmented-survey, tool-learning, RoTBench}. 
Illustrated in Figure~\ref{fig:tool-learning}, upon receiving a user request, the LLM scrutinizes the user's needs, prompts for sufficient information, selects the appropriate tool, and inputs the required parameters in the specified format. Subsequently, the tool interacts with the environment to furnish feedback to the LLM. The LLM then employ logical reasoning based on the initial request, iterating through these steps until a conclusive answer is achieved.

Owing to the intricate nature of tool learning, initial evaluations heavily relied on manual efforts, engaging experts to assess the accuracy of LLMs tool invocation~\cite{Toolalpaca}.
Despite its reasonable effectiveness, the manpower costs hinder widespread adoption. Currently, researchers are exploring automated evaluation methods. One aspect is indirectly assessed by analyzing the performance improvement achieved through the use of tools in downstream tasks~\cite{Toolformer, ToolQA}, while the other is directly evaluated by formulating rules to measure the \emph{exact match} between the tools chosen by LLMs and the expected results~\cite{metatool}.

However, these methods suffer from two significant drawbacks. One constraint lies in their limited applicability, primarily applicable to scenarios where tools can be predefined. Given the similarity among different tools (e.g., the ability of various search software to process the same query) and the variability in information provided by the same tool at different times (e.g., real-time updates of weather information), these methods struggle to capture the complexity of real-world applications involving diverse tools.
Another limitation is their exclusive focus on evaluating the outcomes of tool selection, neglecting the intricate capabilities required for LLMs to use tools. Tool learning involves more than merely selecting a tool; it integrates the LLMs capabilities in comprehending instructions, logical reasoning, and generalizing information. Therefore, there is a necessity for a thorough examination of how various capabilities influence the entire process of tool learning.

To fill this gap, we introduce \emph{ToolEyes}, a fine-grained system tailored for the evaluation of LLMs' tool learning capabilities in real-world scenarios.\footnote{Detailed comparison of ToolEyes with existing benchmarks can be found in Appendix~\ref{sec:comparison}.}
The system meticulously formulates seven authentic scenarios, covering text generation, data understanding, real-time search, application manipulation, personal life, information retrieval, and financial transactions.
Simultaneously, ToolEyes centers its attention on five essential capabilities vital to the tool learning for LLMs: \emph{format alignment}, \emph{intent comprehension}, \emph{behavior planning}, \emph{tool selection}, and \emph{answer organization}.
Moreover, the system establishes a tool library comprising 568 tools, serving as an interface for LLMs to interact with the environment.

We evaluate ten LLMs across three sources (i.e., open-source, tool-oriented, and closed-source), and identify scenario preferences and constrained cognitive capabilities in tool learning. Notably, augmenting model parameters exacerbates the impairment of tool learning performance.

The main contributions of our work are summarized as follows:
\begin{itemize}
    \item We propose \emph{ToolEyes}, a fine-grained system for the evaluation of LLMs' tool learning capabilities, containing seven diverse real-world scenarios and 568 tools.
    \item We perform an in-depth analysis of the capabilities required for LLMs to effectively engage in tool learning across five dimensions, providing a comprehensive examination of the intricate tool learning process.
    \item We evaluate ten LLMs across three categories and discover their inclination toward specific scenarios and restricted cognitive abilities. These findings provide instructive insights for the future development of tool learning.
\end{itemize}

\begin{figure*}[!t]
    \centering
    \includegraphics[width=\linewidth]{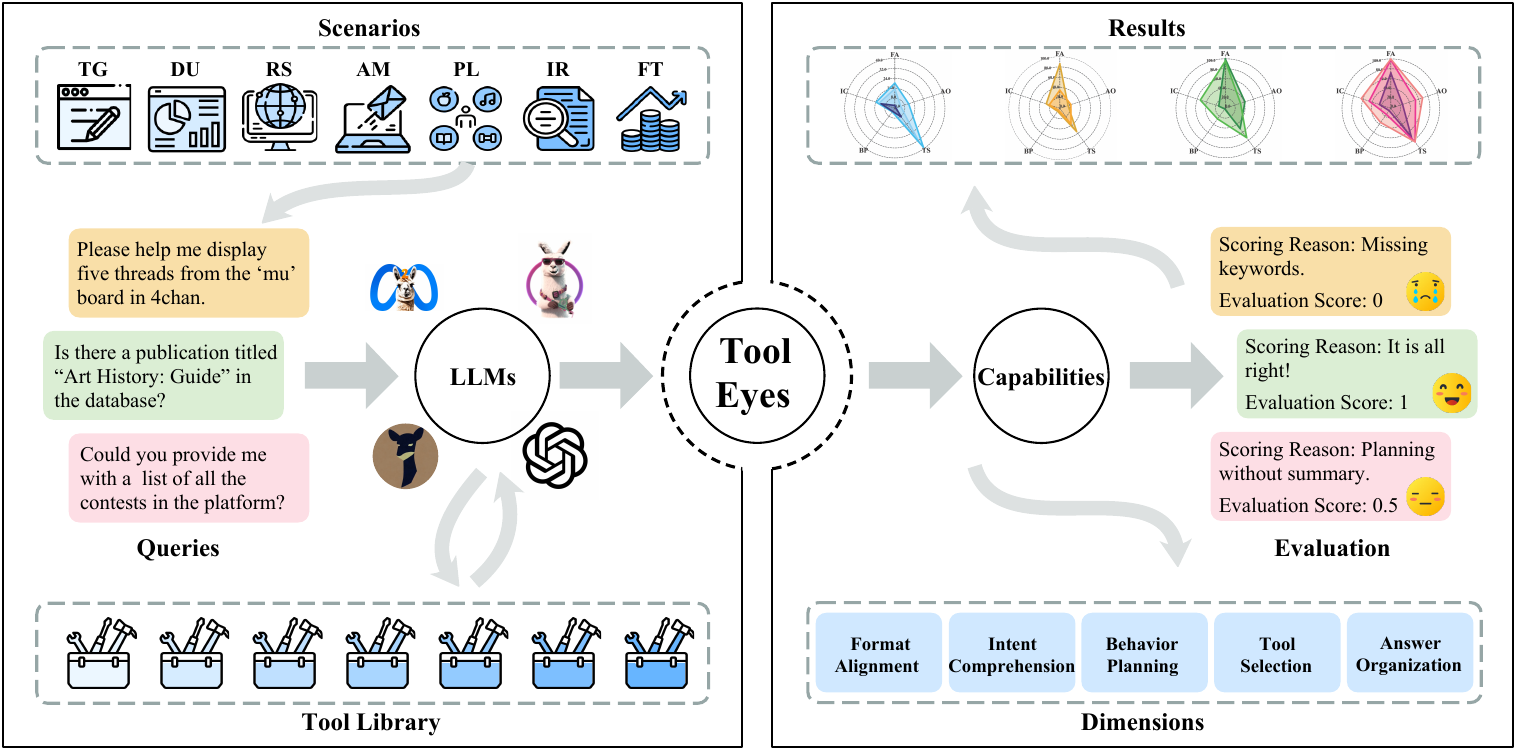}
    \caption{The framework of ToolEyes. ToolEyes formulates seven distinct real-world scenarios. Each scenario incorporates a collection of related tools that LLMs can utilize to engage with the physical world and meet users' practical needs. By evaluating LLMs' capabilities across five dimensions, the system proficiently oversees the entirety of the tool learning process.}
    % \vspace{-4mm}
    \label{fig:ToolEyes}
\end{figure*}

\section{Evaluation System}
As illustrated in Figure~\ref{fig:ToolEyes}, ToolEyes formulates seven distinct real-world scenarios to comprehensively examine the entire tool learning process in accordance with actual application requirements. Each scenario incorporates a collection of related tools that LLMs can utilize to engage with the physical world and meet users' practical needs. By evaluating LLMs' capabilities across five dimensions, the system proficiently oversees the entirety of the tool learning process. 

\subsection{Scenario Construction}

To extend the application of tool learning to capture the intricacies of the physical world, we have devised seven real-world scenarios.

% \paragraph{Text Generation (TG)}
\textbf{Text Generation (TG)} stands out as a highly representative generic scenario, tasking LLMs with generating text that meets user needs while adhering to the query's genre, format, word count, and other specifications. Typical user requests for text generation encompass suggestions, jokes, translations, and more.

% \paragraph{Data Understanding (DU)}
\textbf{Data Understanding (DU)} encapsulates a specialized requirement scenario wherein LLMs are tasked with comprehending user-input data and analyzing it across specific dimensions tailored to user needs, including sentiment analysis, relationship prediction, validity verification, and more.

% \paragraph{Real-Time Search (RS)}
\textbf{Real-Time Search (RS)} is extensively employed in the physical world, requiring LLMs to employ a variety of search tools for gathering information relevant to the user's needs.  Subsequently, LLMs are responsible for compiling and presenting the collected data back to the user in the form of natural language text.

% \paragraph{Application Manipulation (AM)}
\textbf{Application Manipulation (AM)} is a specialized scenario, requiring LLMs to select relevant tools based on user requests. It directly impacts the state of the external environment by executing code, manipulating files, and managing communications, thus surpassing the typical limitations of language model capabilities.

% \paragraph{Personal Life (PL)}
\textbf{Personal Life (PL)} encompasses scenarios tied to personal life needs, prompting LLMs to utilize given tools to gather information on entertainment, food, job, and other relevant topics. Subsequently, LLMs synthesize the acquired information to provide users with effective suggestions.

% \paragraph{Information Retrieval (IR)}
\textbf{Information Retrieval (IR)} is a subset of retrieval tasks, requiring LLMs to retrieve pertinent information from extensive existing databases. This distinguishes itself from RS, which prioritizes instantaneous information. Due to the varied retrieval methods supported by each database, LLMs are compelled to access different databases based on specific requirements.

% \paragraph{Financial Transactions (FT)}
\textbf{Financial Transactions (FT)} includes scenarios that require specialized financial and economic knowledge, prompting LLMs to employ tools for obtaining relevant financial information. Subsequently, LLMs analyze this information to solve the user's problem or provide pertinent advice, which may involve discussions on stock movements or exchange rate fluctuations.

\subsection{Tool Library Building}
\label{sec:toolset}

To establish interfaces for LLMs to engage with the environment, we review existing work for tool design~\cite{Toolformer, ToolQA, Toolllm}, gather real tools across various categories relevant to our constructed scenarios.\footnote{\url{https://github.com/langchain-ai/langchain/tree/master/libs/langchain/langchain/tools}, \url{https://serpapi.com/}}
We systematically rectify tool names and adhered to the GPT-4 format for crafting tool documentation,\footnote{\url{https://platform.openai.com/docs/guides}} creating documentation for each gathered tool. 
Following this organization, each scenario is equipped with a related set of tools, where different tools may serve similar functions.\footnote{Detials for tool collection can be found in Appendix~\ref{sec:tool-collection}.} After aggregation, a comprehensive tool library is established, encompassing 41 categories, 95 subcategories, and 568 tools, capable of fulfilling diverse societal needs. LLMs can invoke these tools using the specified format and retrieve actual information from them.\footnote{Detailed information on tool categories and subcatgories in ecah scenario is provided in the appendix~\ref{sec:detail-tool}.}

\subsection{Human-Driven Data Generation}

Tailored to the constructed scenarios, we engage with a diverse group of professionals linked to each scenario, soliciting their input to identify actual requirements by reviewing the tool documentation. To ensure comprehensive coverage of requirements, we concentrate on one tool subcategory at a time, aiming to encompass the needs of as many tools in that subcategory as possible.\footnote{Specific data generation criteria and examples of data generated for each scenario can be found in Appendix~\ref{sec:detail-data}.} Subsequently, we gathered a total of 382 user queries after thorough manual validation. For a detailed breakdown of the number of tools and queries associated with each scenario, please refer to Table~\ref{tab:data-info}.

\begin{table}[!t]
\centering
\resizebox{\linewidth}{!}{
\begin{tabular}{l cccccccc}
\toprule
\textbf{Scenario} & \textbf{TG} &\textbf{DU} &  \textbf{RS} 
 &\textbf{PL} 
&\textbf{IR}  &\textbf{AM} 
&\textbf{FT} & \textbf{\textit{Total}} \\ \midrule
\textbf{\# Cat} & 5 &5 &  6 
 &
8&9  &
6&2 & 41 \\
\textbf{\# Subcat} & 6 &5 &  14 
 &30&19  &7&14 & 95 \\
\textbf{\# Tool} & 27 &26 &  75 
 &
164&150  &
164&96 & 568 \\
\textbf{\# Query} & 58 &49 &  56  &70&54  &45&50 & 382 \\
\bottomrule
\end{tabular}}
\caption{Statistical information about the data for each scenario. ``\# Cat'' denotes the number of tool categories, ``\# Subcat'' represents the number of tool subcategories, ``\# Tool'' indicates the quantity of tools, and ``\# Query'' represents the number of user queries.}
% \vspace{-4mm}
\label{tab:data-info}
\end{table}

\subsection{LLMs Capability Evaluation}
\label{sec:cap}

Diverging from prior methods that necessitate a predetermined selection of tools, we conduct a comprehensive evaluation of LLMs' interaction with their environments, considering the five dimensions of capability essential for tool learning.

% \paragraph{Format Alignment}
\textbf{Format alignment} stands as a fundamental capability crucial to tool learning, necessitating LLMs to adhere to output formatting requirements in the instructions, ensuring the correct parsing of their output. This includes 1) incorporating corresponding keywords (e.g., Thought, Action, Action Input) to facilitate output separation, and 2) refraining from generating redundant sentences to enable the extraction of tools and parameters.
If the total number of rounds in which LLMs invoke a tool is \(N\), and the number of rounds where the output meets the specified format requirement is \(N_{valid}\), the score \(s_{FA}\) corresponding to its instruction adherence capability is:
\begin{equation}
    s_{FA} = {N_{valid}}/{N}
\end{equation}

% \paragraph{Intent Comprehension}
\textbf{Intent comprehension} hinges on the inherent characteristics of tool learning, focusing on grasping user needs and conducting subsequent analyses. It is crucial to evaluate whether LLMs can continuously update acquired information and adjust solutions to accommodate evolving user input or changing requirements throughout the entire process. To assess this, we determine the intent comprehension capability score for LLMs by evaluating 1) the relevance of their thought processes to user needs and 2) their adaptability to newly provided information during interactions:
\begin{equation}
    s_{IC} \in \left[0, 1\right]
\end{equation}

% \paragraph{Behavior Planning}
\textbf{Behavioral planning} plays a crucial role in facilitating tool learning and assessing the thinking skills of LLMs. Aligned with the insights proposed by~\citet{CoT}, a comprehensive understanding of how LLMs select tools and process information goes beyond mere tool and parameter choices. It is essential for LLMs to concisely summarize relevant information acquired and strategically plan for subsequent steps. When evaluating LLMs' thinking processes, we scrutinize the validity and logical integrity of their thoughts separately. Concerning validity, we obtain the score $s_{b-validity}\in \left[0,1\right]$ by assessing 1) the reasonableness of summarizing the current state, 2) the timeliness of planning for the next sequence of actions, and 3) the diversity of planning. For logical consistency, we calculate the score $s_{b-integrity}\in \left[0,1\right]$ by evaluating 1) grammatical soundness, 2) logical consistency, and 3) the ability to correct thinking. The composite score for behavioral planning capability is determined as follows:
\begin{equation}
    s_{BP} = s_{b-validity} \cdot s_{b-integrity}
\end{equation}

\begin{table*}[!t]
\centering

\resizebox{\linewidth}{!}{
\begin{tabular}{l|l |ccccccc|c} 
\toprule
 \textbf{Source} 
&\textbf{Models}  & \textbf{TG}&  \textbf{DU}&\textbf{RS}& \textbf{AM}&\textbf{PL}&\textbf{IR}& \textbf{FT}& \textbf{\textit{ALL}} \\ \midrule
 \multirow{5}{*}{\textbf{Open-Source}}
&LLaMA-2-chat-7B & 15.33 &24.48 &13.56 &11.45 &12.39 &10.09 &8.33 &13.59 
\\
 
&LLaMA-2-chat-13B & 19.97 &25.06 &15.59 &24.48 &12.62 &15.68 &15.57 &17.98 
\\
 
&LLaMA-2-chat-70B &  3.84 &6.07 &5.77 &9.04 &4.77 &4.03 &4.40 &5.29 
\\ \cmidrule{2-10}
 
&Vicuna-1.5-7B & 51.53 &36.17 &41.10 &32.83 &40.82 &37.42 &27.78 &38.76 
\\
 
&Vicuna-1.5-13B & 25.76 &21.93 &24.02 &32.61 &23.37 &23.00 &20.22 &	24.27 
\\ \midrule
 \multirow{2}{*}{\textbf{Tool-Oriented}} 
&ToolLLaMA-2-7B-v1  & 49.33 &40.85 &40.14 &39.81 &40.56 &40.92 &38.88 &41.61 
\\
 
&ToolLLaMA-2-7B-v2  & 72.90 &54.65 &54.57 &46.49 &58.70 &54.51 &48.00 &56.30 
\\ \midrule
 \multirow{3}{*}{\textbf{Closed-Source}}
&Text-davinvi-003 & 48.56 &48.50 &34.24 &38.68 &34.12 &38.80 &36.65 &39.71 
\\ \cmidrule{2-10}
 
&GPT-3.5-turbo  & 63.25 &60.14 &60.91 &55.06 &61.50 &61.50 &52.86 &59.61 
\\
 &GPT-4  & \textbf{80.24} &  \textbf{71.58} &\textbf{73.99} &  \textbf{70.33}&\textbf{68.06} &\textbf{65.68} & \textbf{61.58} & \textbf{70.31} \\
\bottomrule
\end{tabular}
}
\caption{The performance of the different models in each scenario, tallied in $s_{overall} (\%)$, with ``ALL'' representing their score over all scenarios. The best result in each scenario is \textbf{bolded}.}
% \vspace{-4mm}
\label{tab:main-sce}
\end{table*}

% \paragraph{Tool Selection}
\textbf{Tool selection} is a pivotal aspect of tool learning, assessing the capability to choose suitable tools and input accurate parameters. Recognizing that the approach to problem-solving through tools is not always singular, as seen in the case of querying weather information for two cities, A and B, where querying A first and querying B first are functionally equivalent, we shift away from the previous approach of pre-setting answers. Instead, our emphasis is on authenticity and validity in the process of tool selection.
For the \({i}\)-th round of valid output, our evaluation comprises two key aspects:
1) We scrutinize whether LLMs' tool selection and parameter input align with the requirements. This involves confirming if the selected tool is documented, if the filled parameters correspond to the tool, and if all necessary parameters are included. This assessment is scored in this segment as \(s_{t-reality}^i = 1\) when tool and parameters match the documentation, and 0 otherwise.
2) We prompt LLMs in the instructions to explicitly articulate their thought process behind tool selection, and calculate a match score \(s_{t-match}^i \in \left[0,1\right]\) by comparing their chosen tool with their stated thought. 
Ultimately, the score corresponding to LLMs' tool selection capability is derived as:
\begin{equation}
    s_{TS} = {\sum_{i}s_{t-reality}^i \cdot s_{t-match}^i}/{N_{valid}}
\end{equation}

% \paragraph{Answer Organization}
\textbf{Answer organization} marks the final phase of tool learning, requiring LLMs to amalgamate information gathered throughout the process and furnish a direct response to the user's query. 
This evaluation unfolds in two dimensions:
1) We assess the capability of LLMs to deliver timely responses. Specifically, to safeguard against LLMs entering unproductive quandaries, we define the maximum number of rounds an LLM can engage with the environment for a given query as $N_{max}$. We designate $s_{a-pass}=1$ if the LLM can respond within $N_{max}$ rounds of interactions and 0 otherwise.
2) We scrutinize the quality of responses provided by LLMs. When $s_{a-pass}=1$, the assessment is based on the response's relevance to the user's query and the accuracy of the information conveyed, denoted by $s_{a-quality}$. 
Consequently, the answer organization ability score of an LLM is derived by multiplying these two scores:
\begin{equation}
    s_{AO} = s_{a-pass} \cdot s_{a-quality}
\end{equation}

% \paragraph{Overall}
Upon acquiring the capability scores of LLMs for each of the five dimensions, we establish the overall scores for LLMs' tool learning as:
\begin{equation}
    s_{overall} = \frac{s_{FA}+s_{IC}+s_{BP}+s_{TS}+s_{AO}}{5}
\end{equation}

\begin{table}[!t]
    \centering
    \resizebox{\linewidth}{!}{
    \begin{tabular}{p{0.2\linewidth}|l|cc}
    \toprule
     \textbf{Source} & \textbf{Models} & \textbf{F Statistic} & \textbf{P Value} \\ \midrule
   \multirow{5}{0.2\linewidth}{\textbf{Open-Source}}   &  {LLaMA-2-chat-7B} & 5.82 & $8.20 \times 10^{-6}$ \\
  & {LLaMA-2-chat-13B} & 4.87 & $8.27 \times 10^{-5}$ \\
  & {LLaMA-2-chat-70B} & 2.75 & $1.27 \times 10^{-2}$ \\ \cmidrule{2-4}
  & {Vicuna-1.5-7B} & 15.7 & $4.23 \times 10^{-16}$ \\
  & {Vicuna-1.5-13B} & 1.78 & $1.01 \times 10^{-1}$ \\ \midrule
  \multirow{2}{0.2\linewidth}{\textbf{Tool-Oriented}} & {ToolLLaMA-2-7B-v1} & 10.50 & $8.93 \times 10^{-11}$ \\
& {ToolLLaMA-2-7B-v2} & 14.68 & $4.49 \times 10^{-15}$ \\ \midrule
  \multirow{3}{0.2\linewidth}{\textbf{Closed-Source}} & {Text-davinvi-003} & 7.06 & $3.85 \times 10^{-7}$ \\ \cmidrule{2-4}
  & {GPT-3.5-turbo} & 3.47 & $2.36 \times 10^{-3}$ \\
  & {GPT-4} & 8.47 & $1.23 \times 10^{-8}$ \\
  \bottomrule
    \end{tabular}
    }
    \caption{Welch's ANOVA for $s_{overall}$ across the seven scenarios for various LLMs. A p-value below 0.05 indicate significant differences in the data.}
    % \vspace{-4mm}
    \label{tab:Welch}
\end{table}

\section{Experiments}
To comprehensively assess the tool learning capabilities of various LLMs, we conduct experiments on ten LLMs sourced from three origins, including open-source, tool-oriented, and closed-source.\footnote{The details of the LLMs can be found in Appendix~\ref{sec:model}.}

\subsection{Experimental Setup}

To avoid the effect of unfair testing due to the prompt format during inference, we refer to tool-oriented models and require LLMs to use the {ReAct~\cite{React}} format for output. Since the open-source models were not trained on the tool-learning dataset, we use a five-shot for them and a zero-shot format for all other models.\footnote{The specific prompt can be found in Appendix~\ref{sec:prompt-infer}.}
The maximum allowable interaction turns are set to 9.
It is essential to note that, for all LLMs, our self-constructed tool documentation and user requirements remain out-of-domain.
We set the temperature to 0.3 and top\_p to 0.5 to enhance the diversity of LLMs outputs while ensuring stability.

In the evaluation, $s_{FA}$, $s_{t-reality}$, $s_{a-pass}$ are evaluated based on established rules. Other scores are evaluated by GPT-4.\footnote{The specific prompt can be found in Appendix~\ref{sec:prompt-eval} and analysis for the evaluation quality can be found in Appendix~\ref{sec:ana-eval}.}

\subsection{Results in Different Scenarios}

\begin{figure}[!t]
    \centering
    \includegraphics[width=\linewidth]{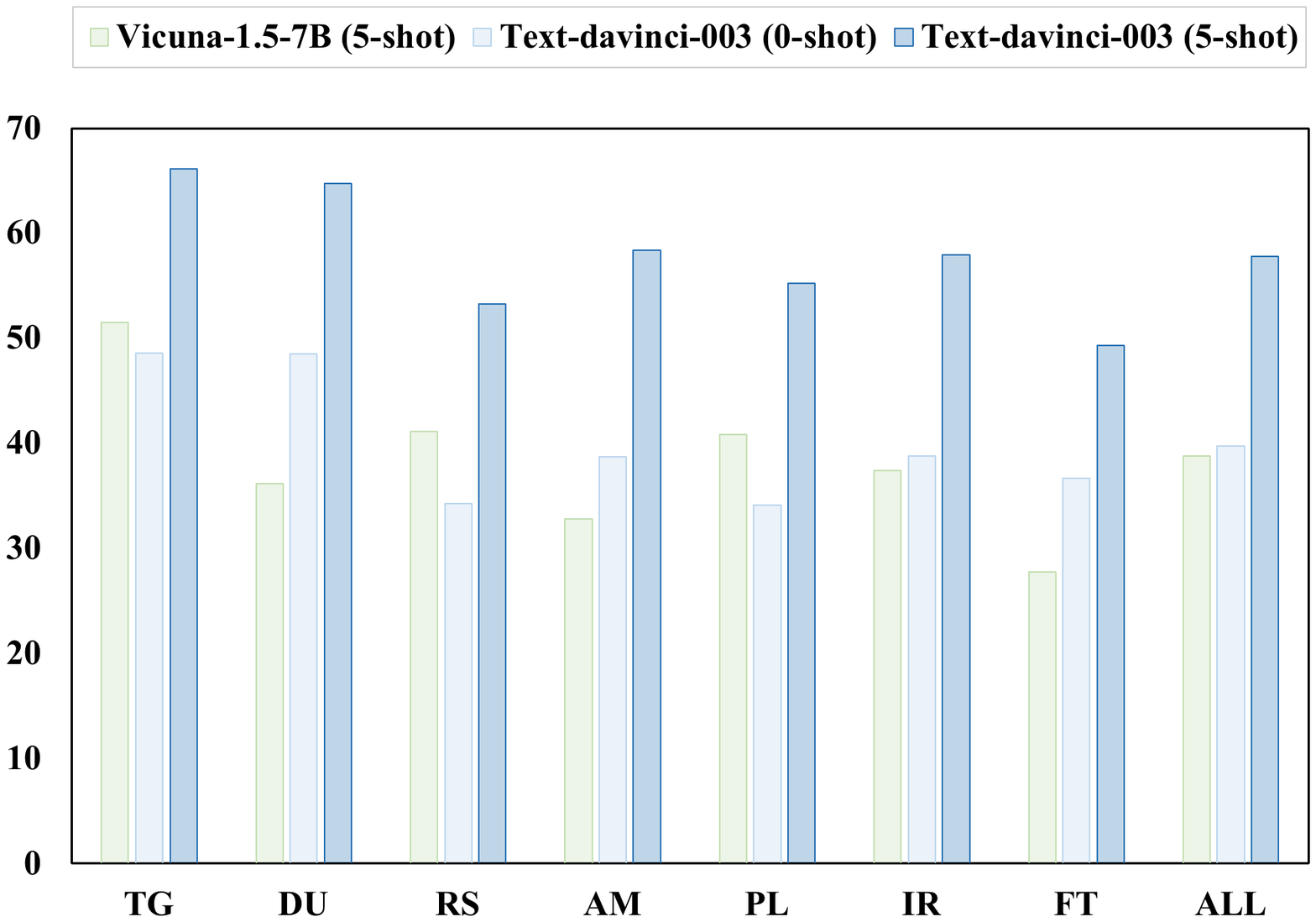}
    \caption{Comparison of the performance of Vicuna-1.5-7B and Text-davinci-003 in each scenario.}
    % \vspace{-4mm}
    \label{fig:shot}
\end{figure}

We evaluate the tool learning performance of the LLMs across seven real-world scenarios, documenting their overall performance scores in Table~\ref{tab:main-sce}.\footnote{Specific capabilities scores for each scenario are available in Appendix~\ref{sec:detail-result}.} There are several interesting observations from the results.

\textbf{LLMs exhibit scenario-specific preferences in tool learning.}
We conduct Welch's ANOVA test~\cite{Welch} to evaluate the performance of each model across seven scenarios. The results in Table~\ref{tab:Welch} unveil noteworthy variations in LLMs performance across these diverse scenarios. Specifically, many LLMs exhibit remarkable proficiency in scenarios such as TG and DU, whereas they demonstrate limitations in scenarios like IR or FT. This discrepancy arises from the fact that, in the former scenarios, the tool's return value can be directly utilized as the final output. In contrast, the return values of tools in the latter scenarios encompass more extraneous information, demanding a heightened ability to generalize relevant information effectively.

\textbf{The variance in tool learning performance between open-source LLMs and closed-source LLMs is considerable.}
Upon evaluating the tool learning capabilities of various source LLMs, closed-source models generally surpass open-source ones, particularly GPT-4. While Vicuna-1.5-7B performs comparably to Text-davinci-003 without demonstrations, Text-davinci-003 surpasses it by 15 points in the five-shot setting (See Figure~\ref{fig:shot}). Moreover, even the leading tool-oriented model ToolLLaMA-2-7B-v2 only achieves 80\% of GPT-4's performance. This underscores a notable opportunity for enhancing tool learning across all categories of LLMs.

\begin{figure}[!t]
    \centering
    \includegraphics[width=\linewidth]{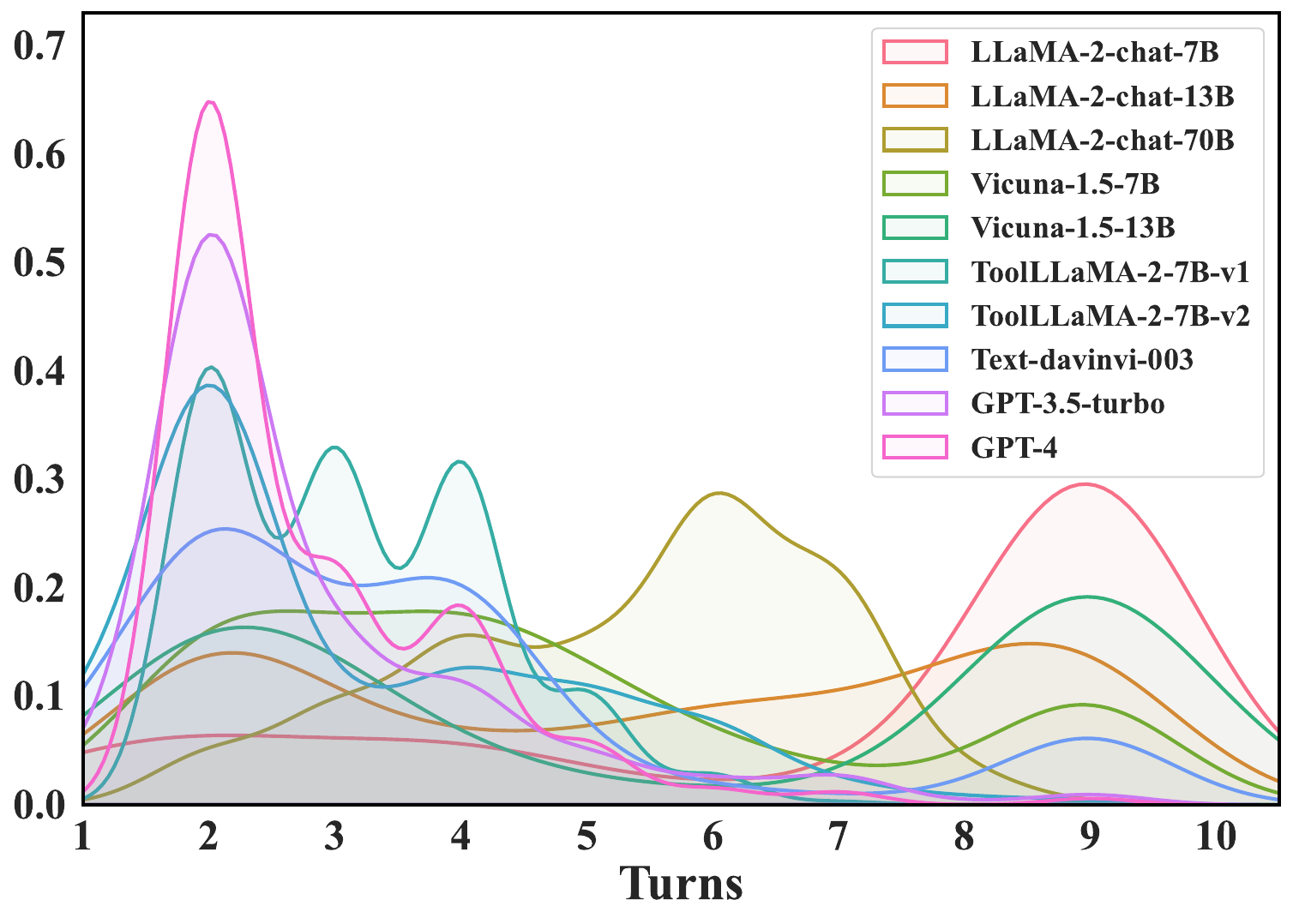}
    \caption{Probability density distribution of the number of turns each LLM interacts with the environment.}
    % \vspace{-4mm}
    \label{fig:turns}
\end{figure}

\textbf{LLMs with superior performance exhibit more effective problem-solving abilities.}
We analyze data across various scenarios to examine the distribution of interaction turns with the environment for different LLMs. The results (Figure~\ref{fig:turns}) demonstrate that, in contrast to open-source LLMs that often necessitate multiple turns to complete tasks, tool-oriented and closed-source LLMs
% , which excel in tool learning tasks, 
can efficiently address problems and meet user needs in a limited number of interaction turns.
On average, LLaMA-2-chat-7B requires 7.0 turns of interaction, a figure significantly higher than the 3.1 turns needed by ToolLLaMA-2-7b-v2 and the 2.8 turns required by GPT-4.

\begin{figure*}[!t]
    \centering
    \includegraphics[width=\linewidth]{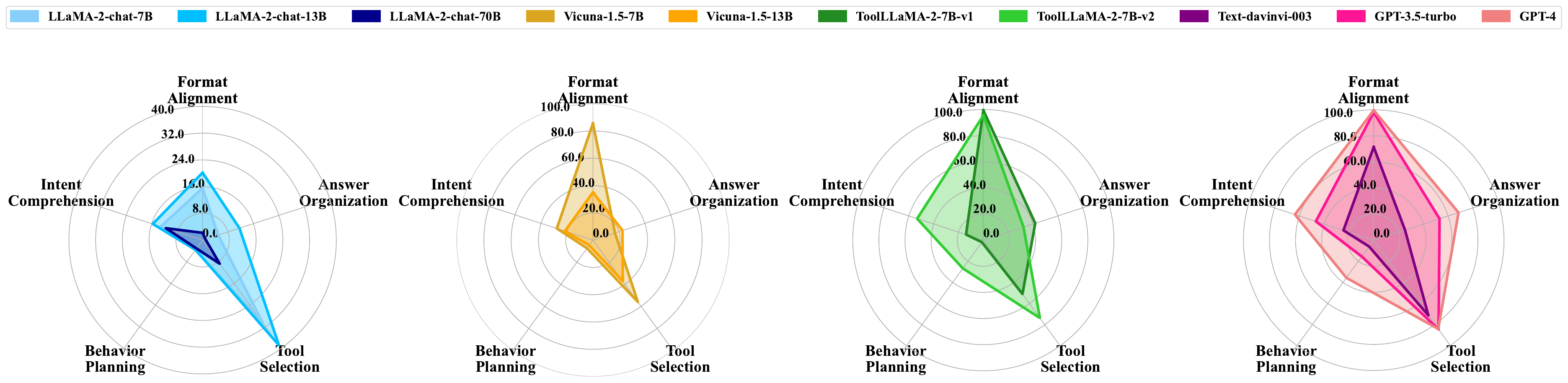}
    \caption{Performance of various LLMs for each capability dimension over all scenarios.}
    % \vspace{-4mm}
    \label{fig:capability}
\end{figure*}

\subsection{Results of Different LLMs Capabilities}

We examine the entirety of the tool learning process, focusing on the five dimensions of capability essential for LLMs to successfully undertake tool learning. The findings, illustrated in Figure~\ref{fig:capability}, unveil noteworthy phenomena that capture our attention.

\textbf{The present constraints in LLMs thinking skills present a substantial obstacle to tool learning}.
Irrespective of their origin, shortcomings in LLMs' behavioral planning skills are apparent across various capabilities essential for effective tool learning. Even the most proficient model, GPT-4, exhibited a mere 35.70\% proficiency in behavioral planning. This underscores a distinct gap in the validity and comprehensiveness of the cognitive processes employed by current LLMs, potentially resulting in suboptimal tool selection, particularly in scenarios demanding multiple interactions with the environment.

\textbf{LLMs' tool learning capabilities are influenced by their optimization goals and training data.}
LLaMA-2-chat-7B, trained based on the LLaMA-2-base-7B, is optimized for generic conversations and aligned using RLHF. Vicuna-1.5-7B prioritizes instruction adherence, relying on a high-quality dataset of SFT instructions for fine-tuning. In contrast, ToolLLaMA-2-7B-v2 is tailored for tool learning and utilizes domain datasets for fine-tuning. Consequently, Vicuna-1.5-7B demonstrates a 73.1\% improvement in format alignment capability compared to LLaMA-2-chat-7B, but its overall performance is still 17.5\% inferior to ToolLLaMA-2-7B-v2.
Meanwhile, in a comparison with ToolLLaMA-2-7B-v1, the training set of ToolLLaMA-2-7B-v2 is optimized for the cognitive processes of LLMs. This optimization significantly enhances tool learning performance, particularly in intent comprehension and behavior planning.

\begin{figure}[!t]
    \centering
    \includegraphics[width=\linewidth]{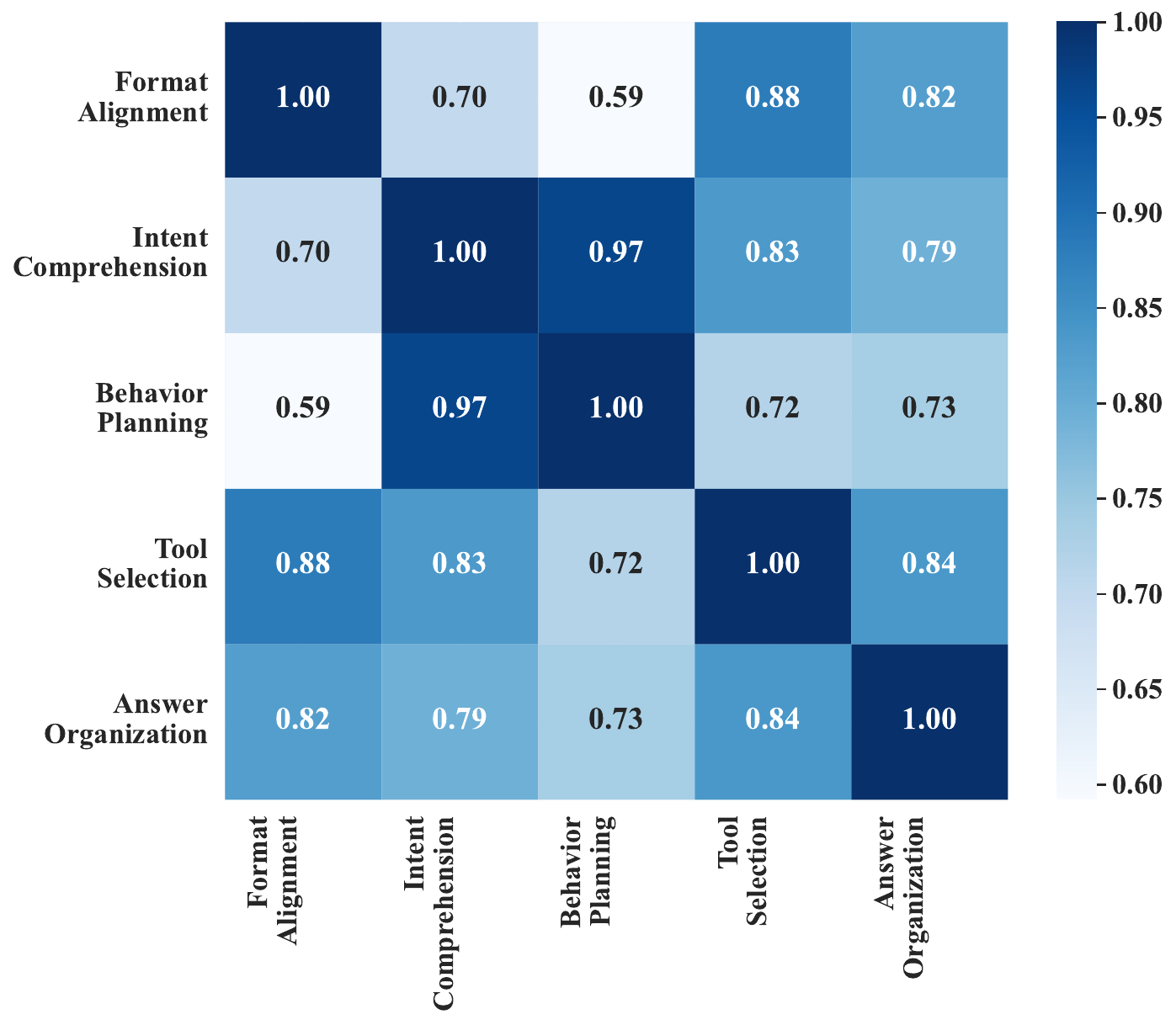}
    \caption{Pearson correlation coefficients between various capabilities dimensions of LLMs.}
    % \vspace{-4mm}
    \label{fig:correlation}
\end{figure}

\textbf{The process of tool learning entails the interaction of various LLMs capabilities.}
We scrutinize the performance across the five capability dimensions and calculate Pearson correlation coefficients, as depicted in Figure~\ref{fig:correlation}.
The analysis uncovers a positive correlation among most LLM competencies. For instance, the correlation between intent comprehension and behavior planning is 0.97, suggesting that LLMs adept at understanding user intent also excel in rational planning. Additionally, correlations surpassing 0.7 are observed between LLMs' tool selection and other capabilities. This underscores that tool learning is a multifaceted process requiring the synergy of multiple capabilities. Therefore, evaluating tool learning should extend beyond assessing tool selection outcomes.

\subsection{Why do LLMs Capabilities NOT Increase with Size?}

In contrast to prior studies that suggest increasing model parameters enhances the capabilities of LLMs~\cite{scale_1, scale_2, emergent}, our findings, depicted in Table~\ref{tab:main-sce} and Figure~\ref{fig:capability}, reveal a noteworthy phenomenon. As the model size increases, there appears to be a potential weakening of the instrumental learning capabilities within the LLaMA-2-chat and Vicuna-1.5 family of models.
To illuminate this phenomenon, we conduct a thorough analysis of model performance. Our study discerns that these limitations arise from inherent behavioral characteristics of LLMs.\footnote{Some typical examples can be found in Appendix~\ref{sec:error}.}

\begin{figure}[!t]
    \centering
    \includegraphics[width=\linewidth]{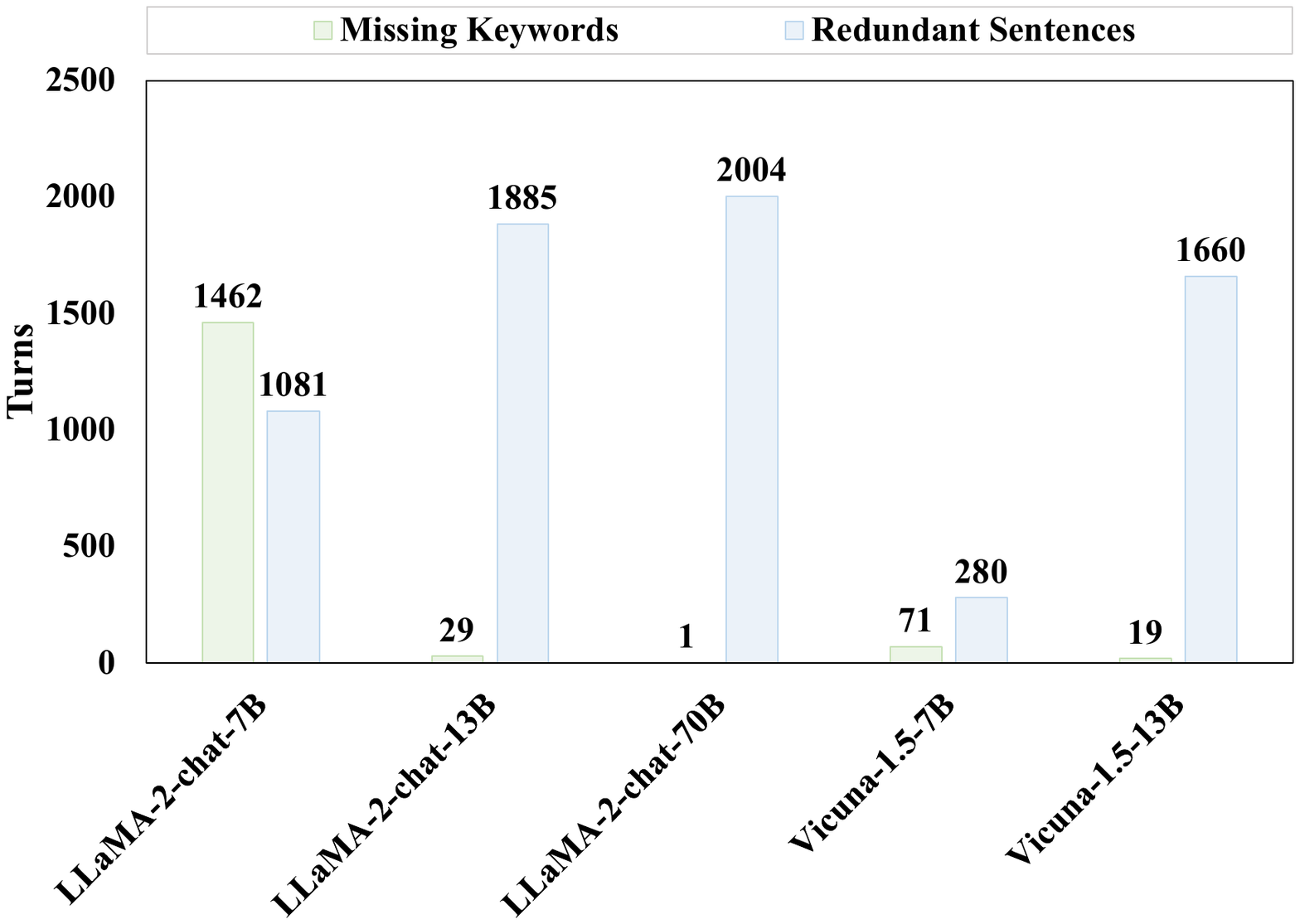}
    \caption{Turns with missing keywords and turns with redundant sentences in LLMs output.}
    % \vspace{-4mm}
    \label{fig:FA-detail}
\end{figure}

\textbf{Aligning with dialog prompts LLMs to generate redundant sentences.}
As explained in Section~\ref{sec:cap}, format alignment entails producing specified keywords while minimizing redundancy. We quantify instances of these errors across all scenarios for the LLaMA-2-chat and Vicuna-1.5 family of models. The results in Figure~\ref{fig:FA-detail} depict a notable increase in the number of turns featuring redundant sentences as the number of parameters increases. This phenomenon can be attributed to LLMs appending extra sentences at the end of tool selection to align more closely with everyday conversations. This behavior is particularly evident in models trained on conversational data, and the impact is magnified with larger parameter sizes. Consequently, interactions by LLaMA-2-chat-70B fail completely in 91\% of the test data, resulting in its markedly poor overall performance.

\textbf{The automatic generation of escaped characters in Vicuna-1.5 leads to tool selection hallucinations.}
To examine the disparity in tool selection performance between Vicuna-1.5-13B and Vicuna-1.5-7B, we compute the average scores of $s_{t-reality}$ and $s_{t-match}$ for both models across all scenarios. The findings in Table~\ref{tab:tool-selection} highlight that the primary factor contributing to the diminished tool selection capability in Vicuna-1.5-13B is a more pronounced issue with tool selection hallucinations. This issue arises from the automatic inclusion of redundant escape characters by Vicuna-1.5, resulting in tool and parameter names that do not align with the information in the tool library. The exacerbation of this phenomenon in Vicuna-1.5-13B is attributed to its utilization of a larger training corpus. 

It is noteworthy that LLaMA-2-chat-13B exhibits markedly improved answer organization compared to LLaMA-2-chat-7B. This is attributed to the tendency of LLaMA-2-chat-7B's responses to deviate from the user's query, leading to a significant decline in quality.
% An instance of this is illustrated in row 6 of Table~\ref{tab:examples}. 
Consequently, as the number of parameters increases, the model's core abilities are enhanced. However, concurrently, its behavioral characteristics, which deviate from the task requirements, are amplified, thereby impacting the overall performance of the model.

\section{Related Works}

\paragraph{Tool Learning}
Since LLMs exhibit the ability to reason and make decisions in intricate interactive environments~\cite{WebGPT}, researchers are keen to harness their potential in addressing more complex social needs through the integration of external tools.
Currently, LLMs' tool learning can be specifically classified into two categories: tool-oriented learning and tool-augmented learning. The former concentrates on enhancing the model's ability to use tools, emphasizing the training of LLMs to become tool experts through specific techniques~\cite{Toolken, oriented_1, TPTU}. The latter, on the other hand, focuses on task processing, where tools are provided as a non-essential means for LLMs to handle tasks~\cite{Tokens, Chamelon, RestGPT}. 
In both scenarios, LLMs' tool learning entails the integration of understanding instructions, logical reasoning, and generalizing information. 
In this paper, we evaluate the five capabilities required by LLMs and analyze the intricate process of tool learning.

\begin{table}[!t]
    \centering
    % \resizebox{\linewidth}{!}{
    \begin{tabular}{lcc}
    \toprule
         Models& $s_{t-reality}$& $s_{t-match}$\\ \midrule
         Vicuna-1.5-7B&  \textbf{63.49}& 89.32\\
         Vicuna-1.5-13B&  51.86& \textbf{93.14}\\
         \bottomrule
    \end{tabular}
    % }
    \caption{$s_{t-reality}$ and $s_{t-match}$ (\%) of Vicuna-1.5.}
    % \vspace{-4mm}
    \label{tab:tool-selection}
\end{table}

\paragraph{Evaluations for Tool Learning}
Existing tool learning evaluations can be broadly classified into three pathways. The first involves manual reviews~\cite{Toolalpaca}, wherein experts familiar with the tool analyze each step of LLMs tool learning to identify problem areas. While effective, the high cost of manpower and time poses challenges for practical application. The second pathway compares the performance of LLMs in downstream tasks before and after utilizing tools, aiming to assess their ability~\cite{GeneGPT, Toolformer, ToolQA}. However, this method relies on tool-task correlations and lacks generalizability to large-scale tool libraries. The recommended approach is to establish scenarios for automated evaluation, but the current practice demands predefined identification of LLMs tool selection and responses, limiting adaptability to real-world environments~\cite{GPT4tool, APIBank, metatool}. To address these limitations, we introduce a fine-grained tool learning evaluation system, enabling in-depth analysis across five capability dimensions throughout the entire tool learning process in the real-world scenarios.

\section{Conclusion}
In this paper, we introduce ToolEyes, a system designed for the fine-grained evaluation of LLMs' tool learning capabilities. The system encompasses 600 tools whose performance undergoes evaluation in seven real-world scenarios across five capability dimensions, spanning the entirety of the tool learning process. The evaluation outcomes include ten different LLMs span three categories, offering valuable insights to inform the ongoing development of tool learning.

% % \clearpage
% \newpage

\section*{Limitations}
While we have established a fine-grained tool learning evaluation system, conducted a comprehensive analysis of commonly used LLMs for tool learning, and outlined directions for future research, our work possesses two notable limitations.
Firstly, we have not developed a novel LLM dedicated to tool learning, aiming to overcome the current deficiencies in tool learning capabilities exhibited by existing LLMs. On a positive note, we have identified key avenues for improvement, which will guide our forthcoming research endeavors.
Secondly, the cost associated with scoring using GPT-4 limited our ability to evaluate all existing LLMs. It i important to highlight that we carefully choose the most representative LLMs from each source for analyzing, aiming to capture the overall problem. Additionally, we plan to explore the possibility of gathering more data to develop a dedicated scoring model, with the intention of mitigating future expenses.

\section*{Acknowledgements}
The authors wish to thank the anonymous reviewers for their helpful comments. This work was partially funded by the Major Key Project of PCL under Grant PCL2024A06, National Natural Science Foundation of China (No. 62476061,62206057,62076069), Shanghai Rising-Star Program (23QA1400200), Natural Science Foundation of Shanghai (23ZR1403500), Program of Shanghai Academic Research Leader under grant 22XD1401100.

\bibliography{custom}

% \clearpage
\appendix

\section{Comparison of ToolEyes with Existing Benchmarks}
\label{sec:comparison}
As described in Section~\ref{sec:intro}, currently available tool learning assessment schemes either have a limited scope of application or a limited focus on dimensionality. To illustrate this, we compare ToolEyes with existing tool learning assessment methods in Table~\ref{tab:comparison}.
As shown, ToolEyes overcomes the shortcomings of existing benchmarks, enabling a fine-grained and comprehensive evaluation.

\begin{table*}[h]
\centering
\resizebox{\linewidth}{!}
{
\begin{tabular}{lcccccccccc}
\toprule
\multirow{2}*{\textbf{Aspect}} & {\textbf{ToolEyes}} & {\textbf{APIBench}} & {\textbf{ToolBench1}} & \textbf{ToolAlpaca}
&\textbf{ToolQA}
& {\textbf{ToolBench2}}& {\textbf{API-Bank}}  &\textbf{MetaTool}
& {\textbf{TaskBench}} & {\textbf{TEval}}  \\
& (Ours)  & \cite{APIBench} & \cite{oriented_1}  & \cite{Toolalpaca} 
&\cite{ToolQA}
& \cite{Toolllm} & \cite{APIBank}  &
\cite{metatool}
& \cite{TaskBench} & \cite{TEval} \\
\midrule

\textbf{Real-world Scenarios} & \textcolor{green!50!black}{\checkmark} & \textcolor{green!50!black}{\checkmark} & \textcolor{red}{\texttimes}  & \textcolor{red}{\texttimes} 
&\textcolor{green!50!black}{\checkmark} 
& \textcolor{green!50!black}{\checkmark} & \textcolor{green!50!black}{\checkmark}  &\textcolor{green!50!black}{\checkmark}
& \textcolor{green!50!black}{\checkmark} & \textcolor{green!50!black}{\checkmark}  \\

\textbf{Manual Crafted Queries} & \textcolor{green!50!black}{\checkmark} & \textcolor{red}{\texttimes} & \textcolor{red}{\texttimes}  & \textcolor{red}{\texttimes} 
&\textcolor{red}{\texttimes} 
& \textcolor{red}{\texttimes} & \textcolor{red}{\texttimes}  &\textcolor{red}{\texttimes}
& \textcolor{red}{\texttimes} & \textcolor{red}{\texttimes}  \\

\textbf{Multi-step Reasoning} & \textcolor{green!50!black}{\checkmark} & \textcolor{red}{\texttimes} & \textcolor{green!50!black}{\checkmark}  & \textcolor{green!50!black}{\checkmark} 
&\textcolor{green!50!black}{\checkmark} 
& \textcolor{green!50!black}{\checkmark} & \textcolor{green!50!black}{\checkmark}  &
\textcolor{red}{\texttimes}
& \textcolor{green!50!black}{\checkmark} & \textcolor{red}{\texttimes}  \\

\textbf{Automatic Evaluation} & \textcolor{green!50!black}{\checkmark} & \textcolor{green!50!black}{\checkmark} & \textcolor{green!50!black}{\checkmark}  & \textcolor{red}{\texttimes} 
&\textcolor{red}{\texttimes} 
& \textcolor{green!50!black}{\checkmark} & \textcolor{red}{\texttimes}  &\textcolor{green!50!black}{\checkmark}
& \textcolor{green!50!black}{\checkmark} & \textcolor{green!50!black}{\checkmark}  \\
\midrule

\textbf{Format Alignment} & \textcolor{green!50!black}{\checkmark} & \textcolor{red}{\texttimes} & \textcolor{red}{\texttimes}  & \textcolor{red}{\texttimes} 
&\textcolor{red}{\texttimes} 
& \textcolor{red}{\texttimes} & \textcolor{red}{\texttimes}  &
\textcolor{red}{\texttimes}
& \textcolor{red}{\texttimes} & \textcolor{green!50!black}{\checkmark}  \\

\textbf{Intent Comprehension} & \textcolor{green!50!black}{\checkmark} & \textcolor{green!50!black}{\checkmark} & \textcolor{green!50!black}{\checkmark}  & \textcolor{green!50!black}{\checkmark} 
&\textcolor{red}{\texttimes} 
& \textcolor{green!50!black}{\checkmark} & \textcolor{green!50!black}{\checkmark}  &\textcolor{green!50!black}{\checkmark}
& \textcolor{green!50!black}{\checkmark} & \textcolor{red}{\texttimes}  \\

\textbf{Behavior Planning} & \textcolor{green!50!black}{\checkmark} & \textcolor{red}{\texttimes} & \textcolor{red}{\texttimes}  & \textcolor{green!50!black}{\checkmark} 
&\textcolor{red}{\texttimes} 
& \textcolor{red}{\texttimes} & \textcolor{green!50!black}{\checkmark}  &
\textcolor{red}{\texttimes}
& \textcolor{green!50!black}{\checkmark} & \textcolor{red}{\texttimes}  \\

\textbf{Tool Selection} & \textcolor{green!50!black}{\checkmark} & \textcolor{green!50!black}{\checkmark} & \textcolor{green!50!black}{\checkmark}  & \textcolor{green!50!black}{\checkmark} 
&\textcolor{red}{\texttimes} 
& \textcolor{green!50!black}{\checkmark} & \textcolor{green!50!black}{\checkmark}  &\textcolor{green!50!black}{\checkmark}
& \textcolor{green!50!black}{\checkmark} & \textcolor{green!50!black}{\checkmark}  \\

\textbf{Answer Organization} & \textcolor{green!50!black}{\checkmark} & \textcolor{red}{\texttimes} & \textcolor{red}{\texttimes}  & \textcolor{green!50!black}{\checkmark} &\textcolor{green!50!black}{\checkmark} & \textcolor{green!50!black}{\checkmark} & \textcolor{green!50!black}{\checkmark}  &
\textcolor{red}{\texttimes}
& \textcolor{red}{\texttimes} & \textcolor{green!50!black}{\checkmark}  \\
\bottomrule
\end{tabular}
}
\caption{Comparison of ToolEyes with existing benchmarks.}
\label{tab:comparison}

\end{table*}

% \clearpage
\section{Analysis of the Quality of ToolEyes}
We rigorously examine ToolEyes' evaluation outcomes for various LLMs to validate its reliability as an evaluation system.

\subsection{Alignment with Human Evaluations}
\label{sec:ana-eval}
In ToolEyes, some scores are calculated directly based on established rules, while others necessitate evaluation by GPT-4. Therefore,  we compare the quality of GPT-4 scores with human evaluations.

\paragraph{Qualitative Analysis}
To illustrate the scoring outcomes generated by GPT-4, we present examples of GPT-4 scoring in Table~\ref{tab:example-gpt-1} and Table~\ref{tab:example-gpt-2}. \textbf{Through these examples, we observe GPT-4's adherence to our specified scoring criteria, offering an objective and comprehensive assessment of the tool learning trajectory. The accompanying scoring rationale effectively assures the validity of our verification process.}

\begin{table*}[h]
    \centering
    \resizebox{\linewidth}{!}
    {
    \begin{tabular}{p{\linewidth}}
  \toprule
 \rowcolor{gray!10} \multicolumn{1}{c}{\textit{Trajectory}} \\
    User: Can you give me some information about the stock whose ticker is `amzn'?\newline
\newline
LLMs: Thought: I need to use the `options\_data' tool to get real-time options and stock data for the stock with ticker `amzn'.\newline
Action: options\_data\newline
Action Input: \{``ticker'': ``amzn''\}\newline
\newline
Tool: ...\newline
\newline
LLMs: Thought: I have successfully retrieved the real-time data for the stock with ticker `amzn'. Now I can analyze the data and provide the information requested.\newline
Action: finish\newline
Action Input: \{``answer'': ``Amazon.com Inc. (AMZN) is a company listed on the NASDAQ stock exchange. The stock is currently trading at \$142.12 per share. The market cap of the company is \$1,468,672,426,882. The stock has a 52-week high of \$145.86 and a 52-week low of \$81.43. The P/E ratio of the stock is 73.26. The stock has a year-to-date change of 69.40\%. The market is currently closed.''\}\\
\midrule

\rowcolor{gray!10} \multicolumn{1}{c}{\textit{Evaluation for $s_{b-validity}$}} \\
Scoring Reason: The thought chain does not provide a comprehensive summary of the relevant information from the previous step. The first turn identifies the correct tool to use (`options\_data') but does not mention the need for an API key, which is a required parameter according to the tool document. The second turn states that the data has been retrieved but does not specify what information will be analyzed or provided to the user, nor does it plan for the next step in detail. Additionally, the second turn does not contribute new information towards resolving the user's query, as it does not specify what data was retrieved or how it will be used to answer the user's question.\newline
\newline
Evaluation Score: 3\newline
\newline
The score is low because the thought chain lacks detail in planning, does not summarize the necessary parameters for using the `options\_data' tool, and does not provide distinct steps that build upon each other to resolve the user's query.\\
\midrule

\rowcolor{gray!10} \multicolumn{1}{c}{\textit{Evaluation for $s_{b-intergrity}$}} \\
Scoring Reason: The thought chain is grammatically correct, with no errors present. It is logically consistent, as the first turn correctly identifies the `options\_data' tool as the appropriate tool for retrieving real-time stock data for the ticker `amzn'. The second turn indicates that the data has been successfully retrieved, which is a logical next step after identifying the correct tool. However, the thought chain does not provide any actual information about the stock `amzn' as requested by the user, nor does it indicate a next step to communicate the retrieved data to the user. There is no detection or correction of logical inconsistencies because the thought chain is too brief and incomplete to demonstrate these aspects.\newline
\newline
Evaluation Score: 5\newline
\newline
The score is in the middle range because while the thought chain is grammatically correct and logically consistent with the steps it includes, it fails to complete the user's request by not providing the actual information about the stock `amzn'. It also does not show any attempt to correct or detect logical inconsistencies, mainly because it ends prematurely without fully addressing the user query.\\
\bottomrule
    \end{tabular}
    }
    \caption{An example for the evaluation of $s_{BP}$ by GPT-4.}
    \label{tab:example-gpt-1}
\end{table*}

\begin{table*}[h]
    \centering
    \resizebox{\linewidth}{!}
    {
    \begin{tabular}{p{\linewidth}}
  \toprule
 \rowcolor{gray!10} \multicolumn{1}{c}{\textit{Trajectory}} \\
User: I'm interested in books related to the fantasy genre, like `Harry Potter'. Can you help me find more?\newline
\newline
LLMs: Thought: First, I need to use the `books\_related\_searches' function to find books related to the fantasy genre, specifically similar to `Harry Potter'.\newline
    Action: books\_related\_searches\newline
    Action Input: \{``q'': ``Harry Potter''\}\newline
\newline
Tool: ...\newline
\newline
LLMs: Thought: The search result returned a series of `Harry Potter' books. However, I need to find other books related to the fantasy genre, similar to `Harry Potter'. I should ask the user for more specific preferences to narrow down the search.\newline
    Action: ask\_to\_user\newline
    Action Input: \{``question'': ``Could you please specify any particular elements or themes you enjoy in `Harry Potter' that you'd like to see in other fantasy books?''\}\newline
\newline
User: I really enjoy the magical school setting and the coming-of-age themes in `Harry Potter'. I also like the elements of mystery and adventure in the series.\newline
\newline
LLMs: Thought: Given the user's preferences for a magical school setting, coming-of-age themes, and elements of mystery and adventure, I need to find books that incorporate these elements. However, the current toolset does not allow for such specific searches. I will use the `books\_related\_searches' function again with a more general term `fantasy' to find related books and then manually select those that seem to match the user's preferences.\newline
    Action: books\_related\_searches\newline
    Action Input: \{``q'': ``fantasy''\}\newline
\newline
Tool: ...\newline
\newline
LLMs: Thought: The search results returned a variety of fantasy books. However, based on the user's preferences, the book ``Grimgar of Fantasy and Ash: Volume 1'' by Ao Jyumonji seems to be a good match. It's a fantasy novel that includes elements of adventure and mystery. I will recommend this book to the user.\newline
    Action: finish\newline
    Action Input: \{``answer'': ``Based on your preferences, I recommend `Grimgar of Fantasy and Ash: Volume 1' by Ao Jyumonji. It's a fantasy novel that includes elements of adventure and mystery, similar to `Harry Potter'. You can find it here.''\}\\
\midrule

\rowcolor{gray!10} \multicolumn{1}{c}{\textit{Evaluation for $s_{IC}$}} \\
Scoring Reason: The thought chain demonstrates a consistent focus on resolving the user query by seeking books related to the fantasy genre, similar to `Harry Potter'. The chain adapts to the user's additional information about preferences for a magical school setting, coming-of-age themes, and elements of mystery and adventure. The thought process includes an attempt to refine the search based on these preferences and concludes with a specific book recommendation that aligns with the user's interests. However, the thought chain could have been more effective if it had used the `ask\_to\_user' tool to clarify or confirm the user's preferences before making a recommendation.\newline
\newline
Evaluation Score: 8\newline
\newline
The score is high because the thought chain is relevant and adapts to the user's additional information. It loses some points because it could have engaged the user further to ensure the recommendation was as accurate as possible.\\
\bottomrule
    \end{tabular}
    }
    \caption{An example for the evaluation of $s_{IC}$ by GPT-4.}
    \label{tab:example-gpt-2}
\end{table*}

\begin{figure*}[h]
    \centering
    \includegraphics[width=\linewidth]{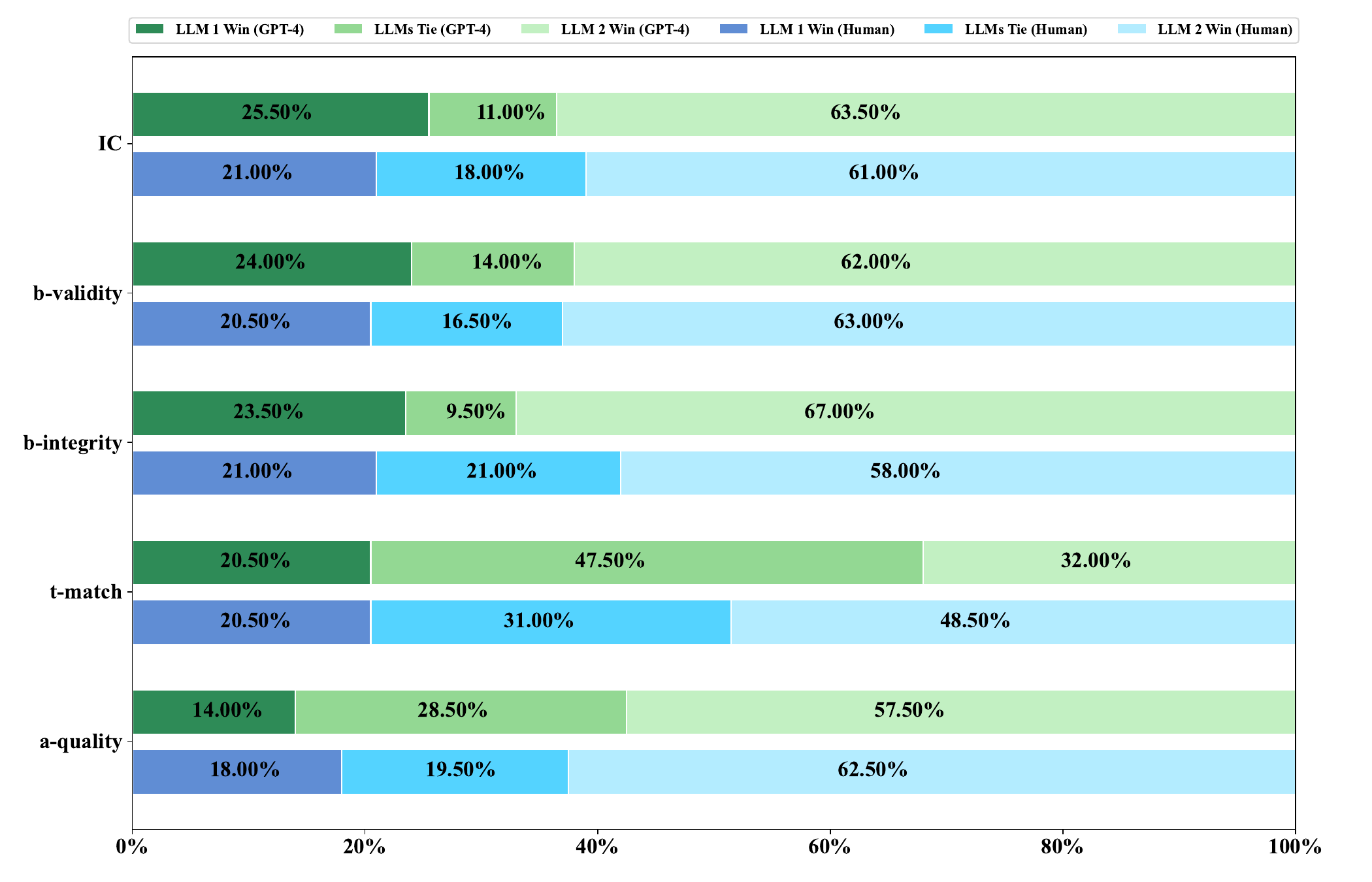}
    \caption{Comparison of GPT-4 and human scoring across various LLMs.}
    \label{fig:human-eval}
\end{figure*}

\paragraph{Quantitative Analysis}
We randomly select 200 sets of tool-learning inference trajectories, each comprising two trajectories from different LLMs for the same user query, facilitating a comparison across various LLM types. Subsequently, we enlist three annotators to evaluate the strengths and weaknesses of these trajectories based on specific metrics outlined in our criteria.\footnote{The inter-annotator agreement score is listed in Table~\ref{tab:agree}.} We then compare the majority of annotation results with those from the GPT-4 evaluation.\footnote{If each of the three annotators provides a different evaluation, we categorize it as a tie between the two trajectories.} As depicted in Figure~\ref{fig:human-eval}, \textbf{the level of agreement in preferences between the GPT-4 evaluation and human evaluation results consistently surpassed 83.50\% across all dimensions, confirming the validity and reliability of our utilization of the GPT-4 assessment.}

\begin{table*}[h]
    \centering
    \begin{tabular}{l|ccccc}
    \toprule
         \textbf{Dimension} & $s_{IC}$ & $s_{b-validity}$ & $s_{b-integrity}$ & $s_{t-match}$ & $s_{a-quality}$\\ \midrule
         \textbf{Kappa}& 0.81 & 0.96 & 0.93 & 0.78 & 0.90 \\ \bottomrule
    \end{tabular}
    \caption{The inter-annotator agreement score of three annotators.}
    \label{tab:agree}
\end{table*}

\begin{figure*}[h]
    \centering
    \includegraphics[width=\linewidth]{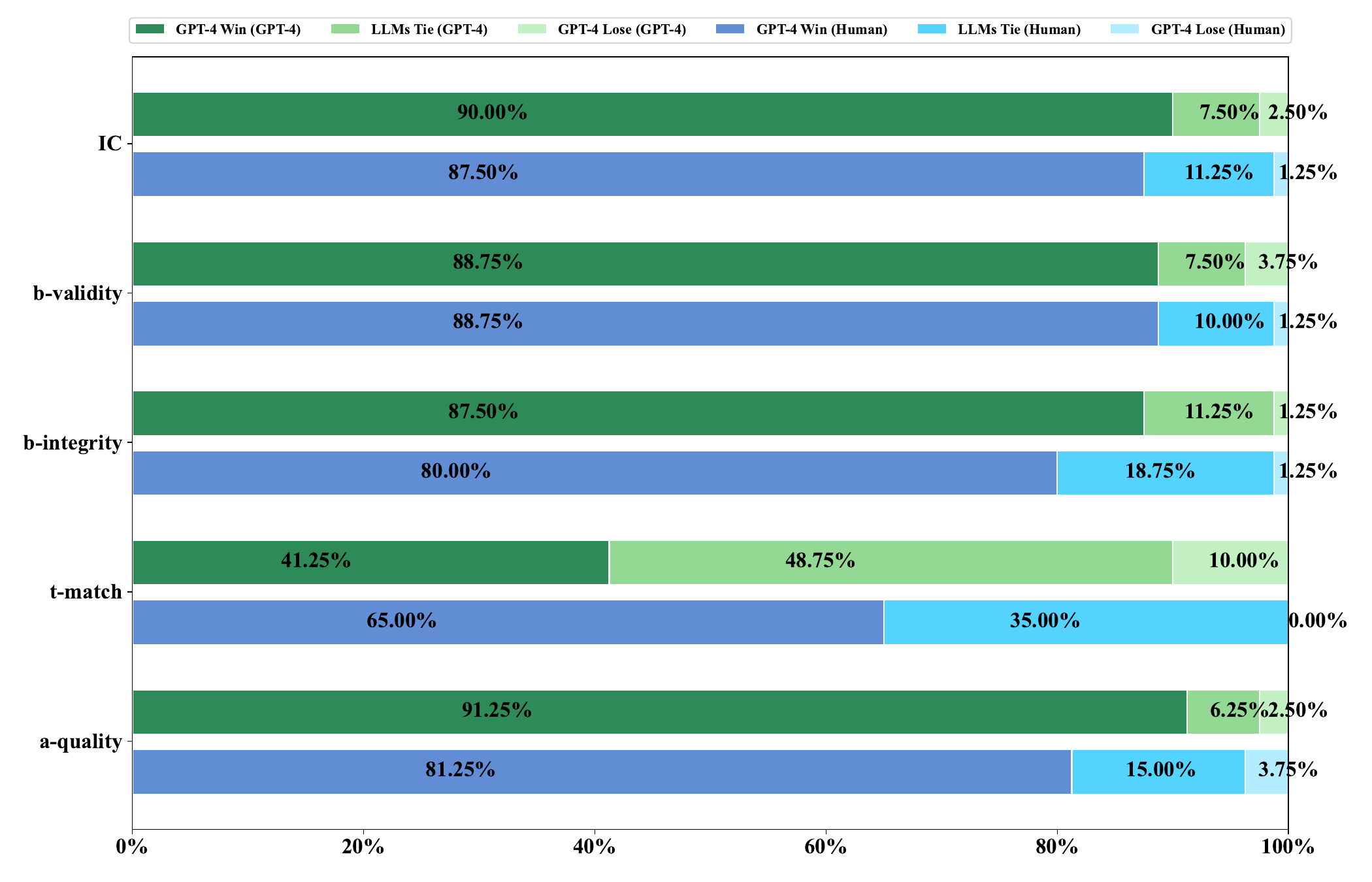}
    \caption{Comparison of GPT-4 and human scoring between GPT-4 and other LLMs.}
    \label{fig:bias}
\end{figure*}

\paragraph{Discussion about Potential Bias}
Using GPT-4 for scoring, even though we validate its consistency with human evaluations, it's crucial to scrutinize whether this scoring method exhibits bias towards GPT-4's own performance.
On one hand, we evaluate the proportion of other LLMs attaining scores equal to or surpassing GPT-4 across various metrics. As shown in Table~\ref{tab:bias}, \textbf{our findings indicate that GPT-4 displays no significant favoritism towards its own performance within the framework of our rubric.}
On the other hand, we examine 80 sets of trajectories between GPT-4 and other LLMs, comparing them with human evaluation outcomes. 
Figure~\ref{fig:bias} indicates sustained high agreement between GPT-4 scores and human scores. Notably, there are marginally lower preferences for GPT-4 results in $s_{t-match}$ compared to human judgments, \textbf{implying the absence of substantial bias towards GPT-4 performance in our assessment program's design.}
The elevated scores attained by GPT-4 can be ascribed to its robust modeling and tool learning capabilities relative to other LLMs. This outcome underscores GPT-4's inherent strengths in these domains rather than any scoring bias.

\begin{table*}[h]
    \centering
    \resizebox{\linewidth}{!}
    {
    \begin{tabular}{l|l|cccc}
    \toprule
    \textbf{Source} & \textbf{Models} & $\mathbf{s_{IC}}$ & $\mathbf{s_{b-validaty}}$ & $\mathbf{s_{b-integrity}}$ & $\mathbf{s_{a-quality}}$ \\ \midrule
    \multirow{5}*{\textbf{Open-Source}} & {LLaMA-2-chat-7B}&11.52&14.40&11.26&6.02
\\
     & {LLaMA-2-chat-13B}&14.40&13.09&13.87&11.26
\\
    & {LLaMA-2-chat-70B}&10.99&13.61&13.87&4.19
\\ \cmidrule{2-6}
     &{Vicuna-1.5-7B}&18.85&19.11&19.11&13.35
\\
     &{Vicuna-1.5-13B}&14.66&15.97&15.18&17.54
\\ \midrule
     \multirow{2}*{\textbf{Tool-Oriented}}&{ToolLLaMA-2-7B-v1}&10.73&11.78&9.95&35.60
\\
     &{ToolLLaMA-2-7B-v2}&46.34&47.64&43.46&21.47
\\ \midrule
     \multirow{2}*{\textbf{Closed-Source}}&{Text-davinci-003}&43.19&33.77&31.68&21.99
\\ \cmidrule{2-6}
     &{GPT-3.5-turbo}&40.31&34.03&36.39&48.43\\
     \bottomrule
    \end{tabular}
    }
    \caption{The proportion of other LLMs achieving scores equal to or higher than GPT-4 across various metrics.}
    \label{tab:bias}
\end{table*}

\subsection{Analysis of Evaluation Metrics}
To ascertain the viability of our proposed five capability dimensions as effective evaluation metrics, we conduct an analysis to evaluate their stability and sensitivity.

\begin{figure*}[ht]
    \centering
    \subfigure[Tool Selection]{
    \includegraphics[width=\linewidth]{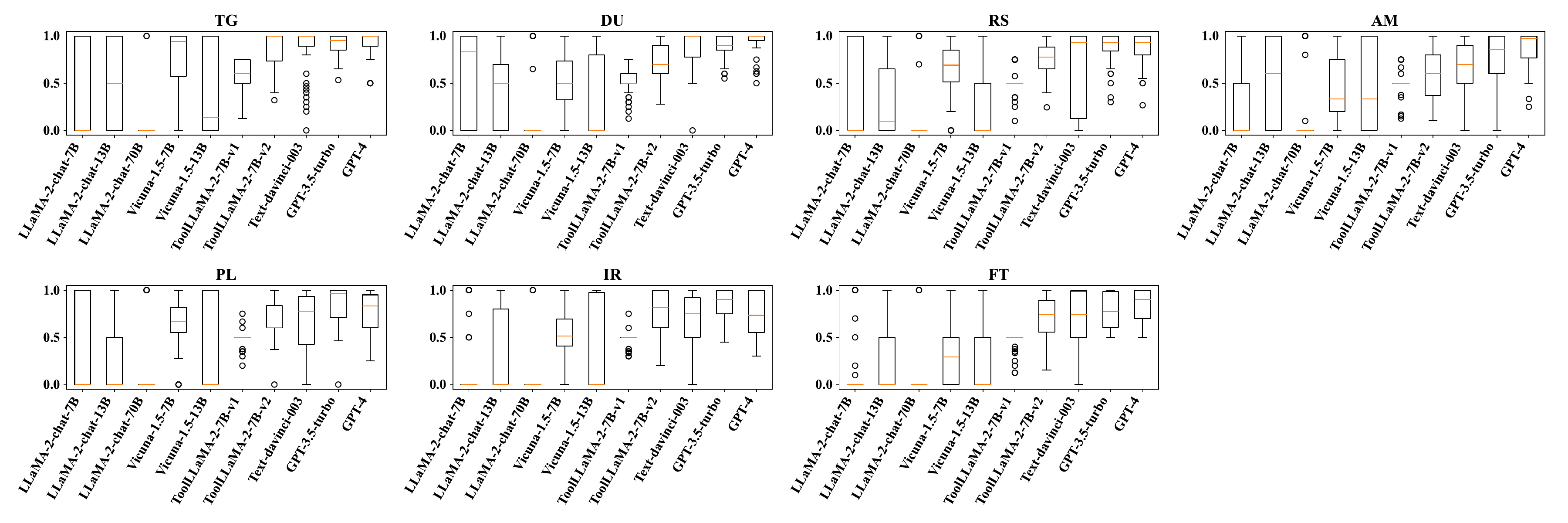}
    }  
    \subfigure[Answer Organization]{
    \includegraphics[width=\linewidth]{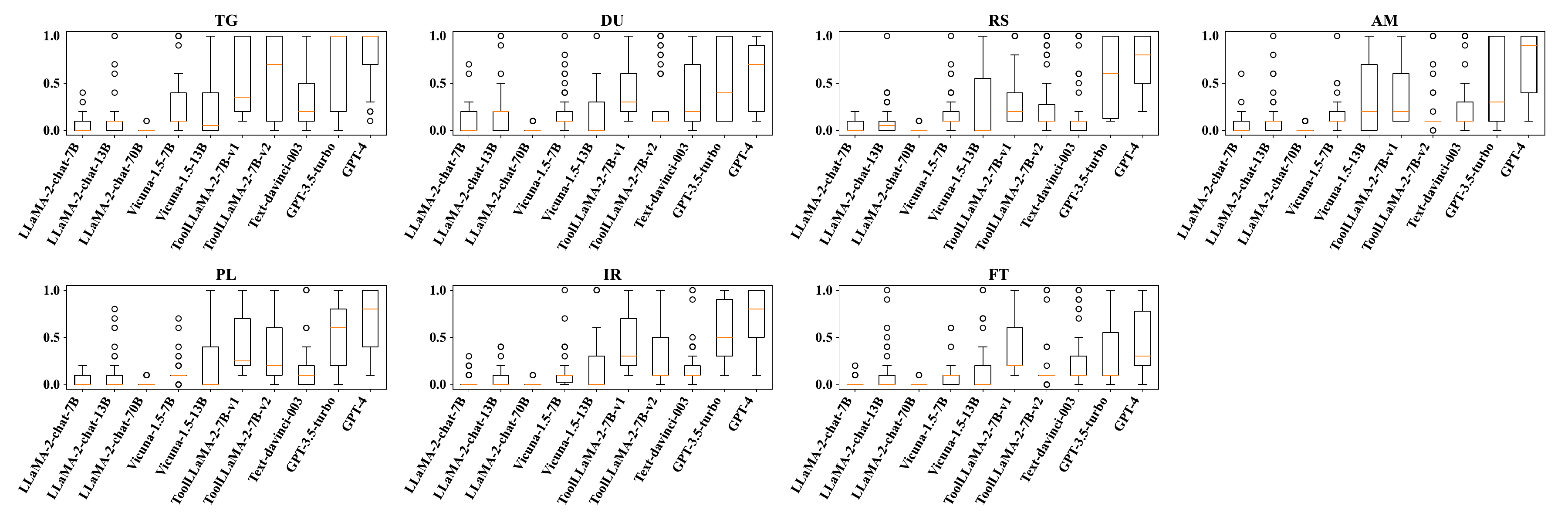}
    }  
    
    \caption{The score distribution of different LLMs in each of the five capability dimensions within each scenario.}
    \label{fig:stable}
\end{figure*}

\paragraph{Stability}
We analyze the score distribution of different LLMs in each of the five capability dimensions within each scenario separately. From the results shown in Figure~\ref{fig:stable}, we find that for the same LLM, the score interval in the same scenario remains fixed for each capability dimension with very little difference. \textbf{This indicates that the metrics we set give similar values for a same LLM in different test samples for the same task.}

\begin{figure*}[ht]
    \centering
    \subfigure[Format Alignment]{
    \includegraphics[width=\linewidth]{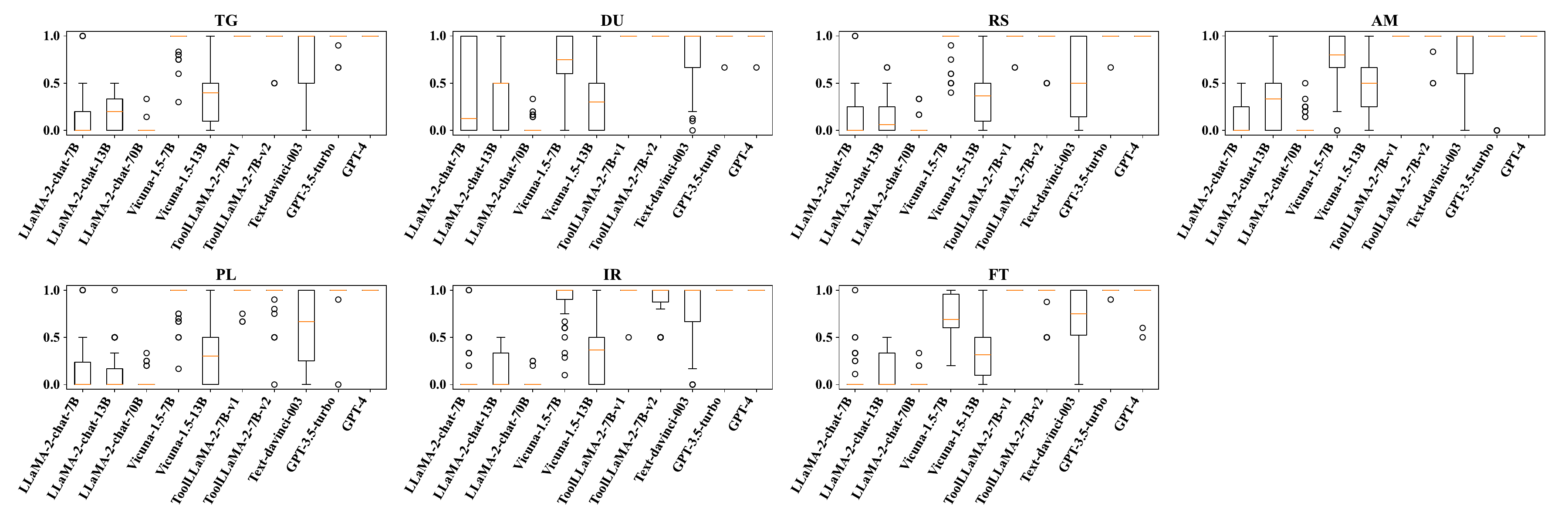}
    }
    \subfigure[Intent Comprehension]{
    \includegraphics[width=\linewidth]{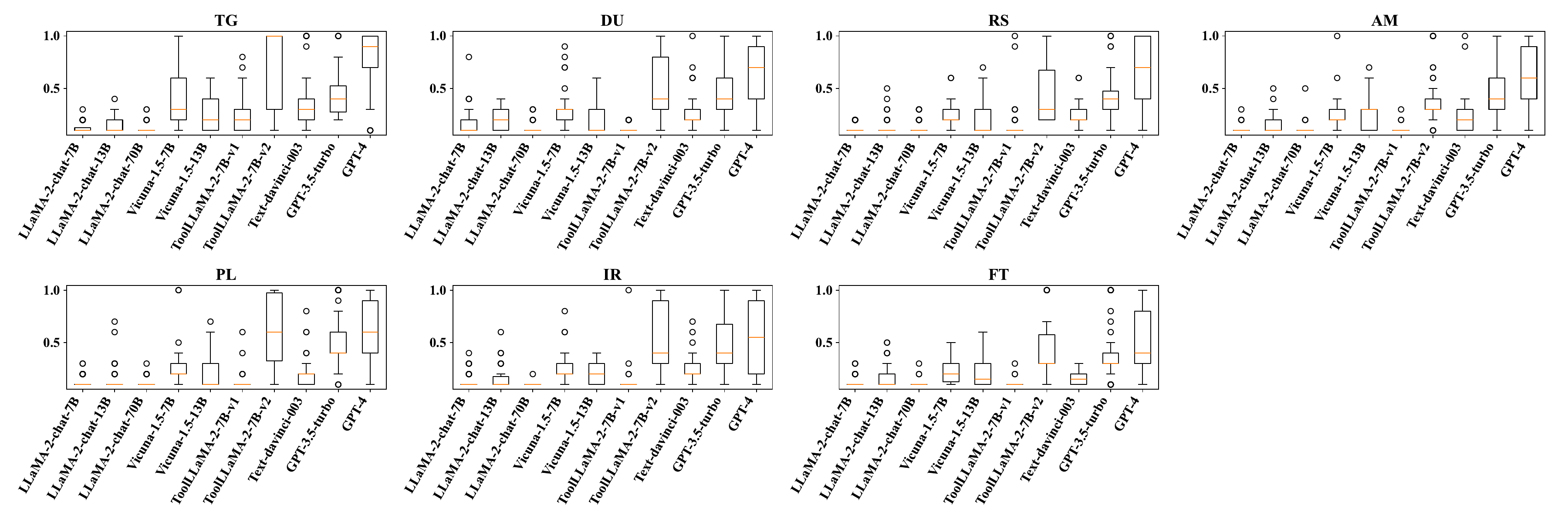}
    }    
\subfigure[Behavior Planning]{
    \includegraphics[width=\linewidth]{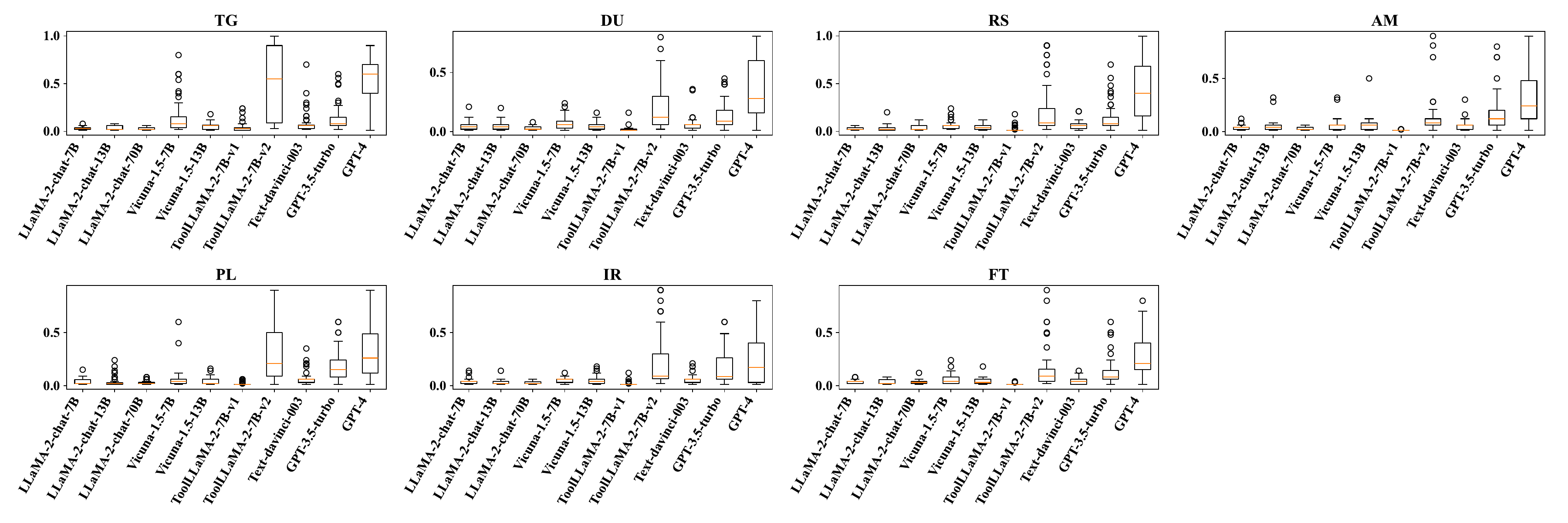}
    }  
    \end{figure*}

\paragraph{Sensitivity}
From Figure~\ref{fig:capability}, Figure~\ref{fig:stable} and Figure~\ref{fig:cap-sce}, it is evident that distinctions in performance across five capability dimensions can be effectively made for different LLMs. For instance, consider ToolLLaMA-2-7B-v1 and ToolLLaMA-2-7B-v2, which share the same base model and training method but differ in model capability. Our evaluation system adeptly discerns variations in their performances across different capability dimensions, aligning well with the training characteristics of LLMs. This differentiation becomes even more pronounced when comparing LLMs from disparate sources. \textbf{Thus, our metrics reliably rank two LLMs, even when their quality differs only slightly.}

% \clearpage
\section{Experimental Details}

\subsection{Details of Tool Collection}
\label{sec:tool-collection}

\paragraph{Criteria for Tool Collection}
To make our collection of tools suitable for the tool learning evaluation, we follow these criteria:

\begin{itemize}
    \item The tools should fit within the seven real-world scenarios constructed and be relevant for daily use.
    \item The tools must be stable and able to be invoked successfully.
    \item The returned data from the tools should not exceed the model's context limits and should be of appropriate length.
    \item The tools should be low-cost to invoke and easily testable by users.
    \item The tools must be well-documented to minimize documentation noise.
\end{itemize}

\paragraph{Process for Tool Collection}
We initially screen a large number of collected tools, excluding those with high costs (more than \$0.50 per call), unstable calls, return values exceeding 4096 tokens, or no return value. We then invite human reviewers to further screen the remaining tools, filtering out those that duplicate functions or are subsets of other tools. Next, we categorize the tools according to the constructed scenarios, filtering out those that do not fit these scenarios and specialized tools (e.g., specialized drawing tools) not useful to general users. Finally, we manually write documentation for the retained tools to ensure the clarity and validity of the tool information.

\subsection{Tool Categories and Subcategories}
\label{sec:detail-tool}

\begin{figure*}[h]
    \centering
    \includegraphics[width=\textwidth]{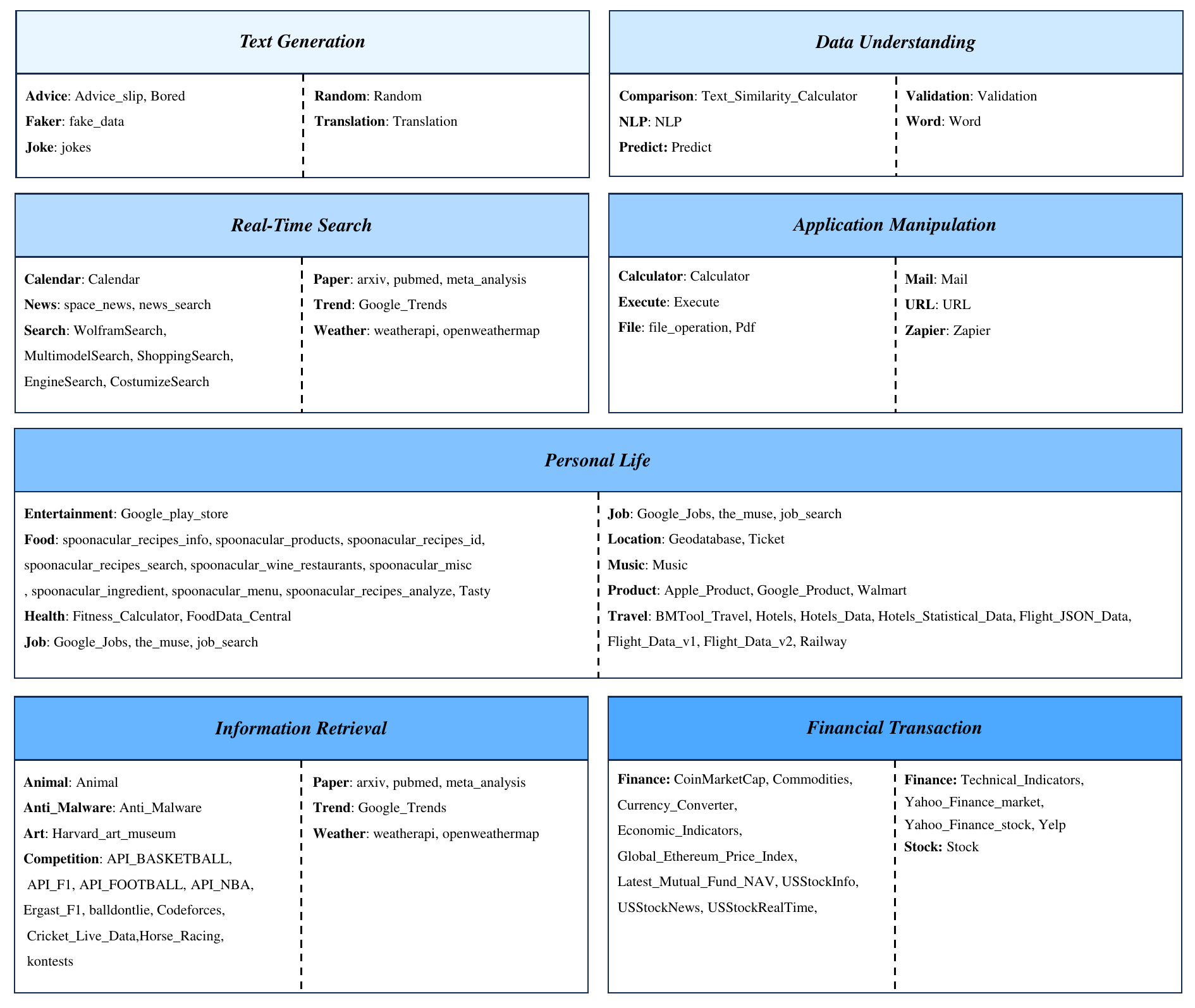}
    \caption{Tool categories and subcategories in each scenario.}
    \label{fig:tool}
\end{figure*}

To establish a connection between LLMs and the environment, we develop a tool library comprising 41 categories and 95 subcategories. The precise names and containment relationships are detailed in Figure~\ref{fig:tool}.

\subsection{Details of Data}
\label{sec:detail-data}

\subsubsection{Criteria for Data Generation}

Professionals related to each scenario are invited to formulate authentic requirements, and the criteria for building these requirements are outlined in Table~\ref{tab:criteria}.

\begin{table*}[!t]
    \centering
    \resizebox{\linewidth}{!}{
    \begin{tabular}{p{\linewidth}}
    \toprule
         As a \{scenario\} professional, your task is to devise pertinent requirements in collaboration with the provided tools, adhering to the following criteria:\\
\\
1. Ensure that the proposed requirements are contextually relevant to your specific scenario and address authentic needs.\\
2. Formulate requirements that are clear, unambiguous, and easily comprehensible.\\
3. Align your requirements with the provided tools, enabling their utilization for acquiring information necessary to address your requirements.\\
4. Your requirements may focus on a single tool or encompass multiple tools simultaneously.\\
5. Cover essential information required for invoking the tool in your requirements, but feel free to omit certain details or rely on common sense.\\
\\
Details about the available tools are provided below:\\
\{Tools\}\\
\bottomrule
    \end{tabular}
    }
    \caption{Guidelines for humans to generate data. ``\{scenario\}'' denotes the scenario name, and ``\{Tools\}'' represents tools within a subcategory.}
    \label{tab:criteria}
\end{table*}

\subsubsection{Examples of Data for Each Scenario}
\begin{table*}[!t]
    \centering
    \resizebox{\linewidth}{!}{
\begin{tabular}{p{\linewidth}}
  \toprule
 \rowcolor{gray!10} \multicolumn{1}{c}{\textit{Text Generation}} \\
    1. How should I say `glass' in Chinese?\\
    2. My friend's wedding is coming up, do you have any advice for the bride?\\
  3. I'm in need of assistance in generating a random string with a length of 8, please give me one.\\
\midrule

\rowcolor{gray!10} \multicolumn{1}{c}{\textit{Data Understanding}} \\ 
  1. Based on their names, what could be the nationalities of John and Maria?\\
2. What emotions are contained in the following text, `Beneath the starry sky, serenity envelops the tranquil meadow, inviting contemplation and inner peace.' \\
3. Please help me assign classes to this text, ``As the gentle waves caress the sandy beach and the sunlight pours down its warm rays, I feel a sense of tranquility and peace within. The beauty and harmony of nature make me forget the hustle and bustle of the city, allowing me to quietly listen to the birds' songs and feel the breath of the wind.''\\
  \midrule

\rowcolor{gray!10} \multicolumn{1}{c}{\textit{Real-Time Search}}  \\
1. Can you tell me what will the weather be like in London for the next week?\\
2. What were the most popular news articles related to technology on August 1st, 2023?\\
3. Can you create a line chart that depicts the search popularity score of restaurant over a period of time?\\
\midrule
  
\rowcolor{gray!10} \multicolumn{1}{c}{\textit{Personal Life}} \\ 
1. What is the distance between Bangkok and Phitsanulok?\\
2. I am looking for films with a style or genre similar to `Pulp Fiction', can you help me find them?\\
3. I will go to Seattle from Beijing next month. Can you make a recommendation on hotels and flight please?\\
\midrule

\rowcolor{gray!10} \multicolumn{1}{c}{\textit{Information Retrieval}} \\
1. Please display five threads from page one of the `mu' board in 4chan.\\
2. Is there a publication titled ``Art History: A Comprehensive Guide'' available at Harvard Art Museum?\\
3. Could you provide me with a comprehensive list of all the contests available on the Codeforces platform?\\
\midrule

\rowcolor{gray!10} \multicolumn{1}{c}{\textit{Application Manipulation}} \\
1. Please summary the content in `./test\_file/read\_test.md' using less than 5 sentences.\\
2. Could you execute this Python expression with Python Interpreter? (123 + 234) / 23 * 19?\\
3. Send an email to xxxxxxxxxx@qq.com with `test\_email' in the subject line and `hello!' in the body.\\
\midrule

 \rowcolor{gray!10} \multicolumn{1}{c}{\textit{Financial Transactions}} \\
1. How much is US GDP these years?\\
2. Show me a summary of the current financial market situation in Germany.\\
3. Please give me most recent daily time series (date, daily open, daily high, daily low, daily close, daily volume) of ``NFLX.''\\
  \bottomrule
\end{tabular}
}
    \caption{Examples of evaluation data in each scenario.}
    \label{tab:details-data}
\end{table*}

Three user queries for each scenario are presented in Table~\ref{tab:details-data}.

\subsection{Model Selection}
\label{sec:model}
To comprehensively assess the tool learning capabilities of various LLMs, we conduct experiments on ten LLMs sourced from three origins, and we will now provide a brief description of each series of models.

\subsubsection{Open-Source LLMs}
\paragraph{LLaMA-2-chat}
LLaMA-2~\cite{LLaMA-2} represents the second iteration of Meta's open-source LLM. Building upon the foundation of LLaMA, it incorporates an increased token count for training and extends the context length to 4096. The LLaMA-2-chat series comprises models fine-tuned for conversational scenarios based on LLaMA-2, employing RLHF~\cite{RLHF} technology for alignment. These models, namely \textbf{LLaMA-2-chat-7B}, \textbf{LLaMA-2-chat-13B}, and \textbf{LLaMA-2-chat-70B}, are distinguished by variations in parameter numbers.

\paragraph{Vicuna-1.5}
Vicuna~\cite{vicuna}, a collection of open-source models introduced by LMSYS, includes Vicuna-1.5, which undergoes fine-tuning from LLaMA-2 using SFT and linear RoPE scaling techniques~\cite{RoPE} . Trained on approximately 125,000 conversations sourced from ShareGPT\footnote{\url{https://sharegpt.com/}}, Vicuna-1.5 exhibits proficient command-following and natural language understanding capabilities.
It is further classified based on model parameter scaling into two specific models: \textbf{Vicuna-1.5-7B} and \textbf{Vicuna-1.5-13B}.

\subsubsection{Tool-Oriented LLMs}
\paragraph{ToolLLaMA-2-7B}
ToolLLaMA~\cite{Toolllm} constitutes a series of specialized LLMs designed for tool learning, developed by Tsinghua University. One notable variant within this series is ToolLLaMA-2-7B, tailored for tool-oriented applications. It is derived from the base model LLaMA-2-7B and fine-tuned using 126 thousand instances of tool learning data associated with 16 thousand APIs through SFT. Depending on the version of the training data employed, it can be further classified into \textbf{ToolLLaMA-2-7B-v1} and \textbf{ToolLLaMA-2-7B-v2}, with the latter showcasing a more advanced thought process in LLMs compared to the former.

\subsubsection{Closed-Source LLMs}
\paragraph{Text-davinci-003}
Text-davinci-003\footnote{\url{https://platform.openai.com/docs/models/gpt-3-5}}, an LLM developed by OpenAI, is part of the GPT-3.5 series designed for tasks that require instruction following. Trained on a combination of text and code data until the fourth quarter of 2021, this model demonstrates proficiency in understanding and generating both natural language and code. With an extensive context window of 16,384 tokens, Text-davinci-003 is fine-tuned for a variety of tasks, including text completion, summarization, and question answering.

\paragraph{GPT-3.5-turbo}
GPT-3.5-turbo\footnote{\url{https://platform.openai.com/docs/models/gpt-3-5}} distinguishes itself as the most powerful and cost-effective model in the GPT-3.5 series. Tailored for chat-based applications, it leverages and enhances the capabilities of Text-davinci-003. This model excels in understanding and generating both natural language and code, while also demonstrating proficiency in traditional text-based tasks.

\paragraph{GPT-4}
GPT-4~\cite{GPT-4} represents OpenAI's cutting-edge system, surpassing its predecessors with the ability to provide safer and more useful responses. Armed with expanded general knowledge and advanced reasoning capabilities, GPT-4 excels in accurately solving puzzles, solidifying its position as one of the most powerful LLMs currently in existence.

\subsection{Details of Result}
\label{sec:detail-result}

\begin{figure*}[!t]
    \centering
    \subfigure[Information Retrieval]{
    
    % \ContinuedFloat
    \includegraphics[width=\linewidth]{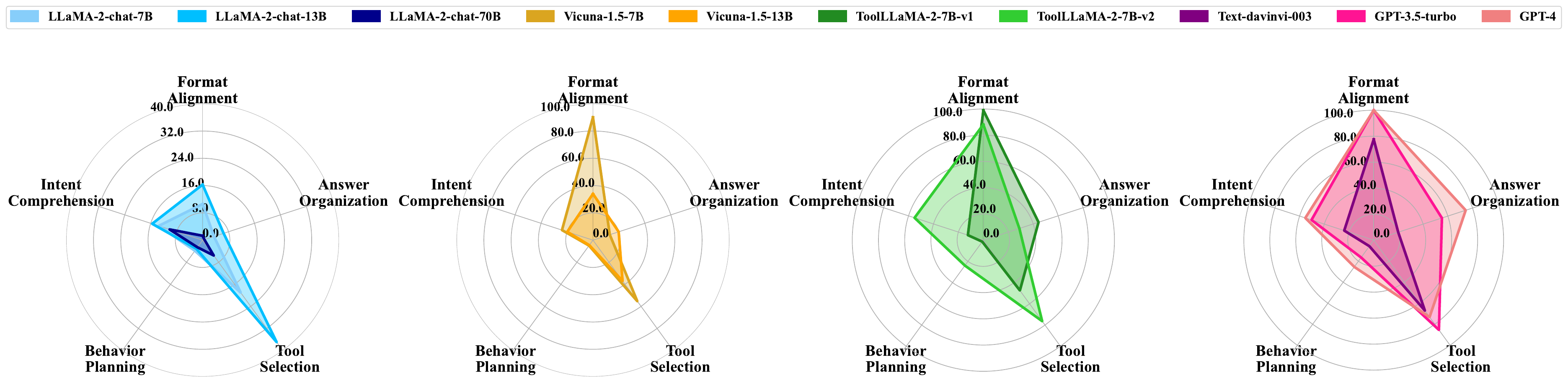}
    }
    \subfigure[Financial Transactions]{
    \includegraphics[width=\linewidth]{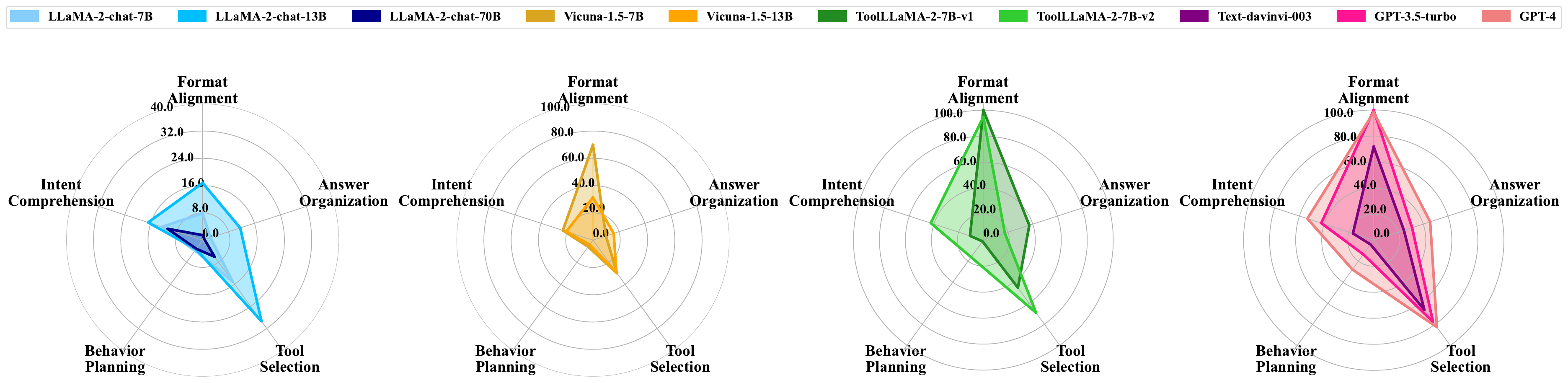}
    }    
    \caption{Performance of various LLMs for each capability dimension in each scenario.}
    \label{fig:cap-sce}
\end{figure*}

We evaluate the capability scores (\%) of the five dimensions of each LLMs in each scenario and plot them in Figure~\ref{fig:cap-sce}.

\begin{figure*}[!t]
    \centering
    \subfigure[Text Generation]{
    \includegraphics[width=\linewidth]{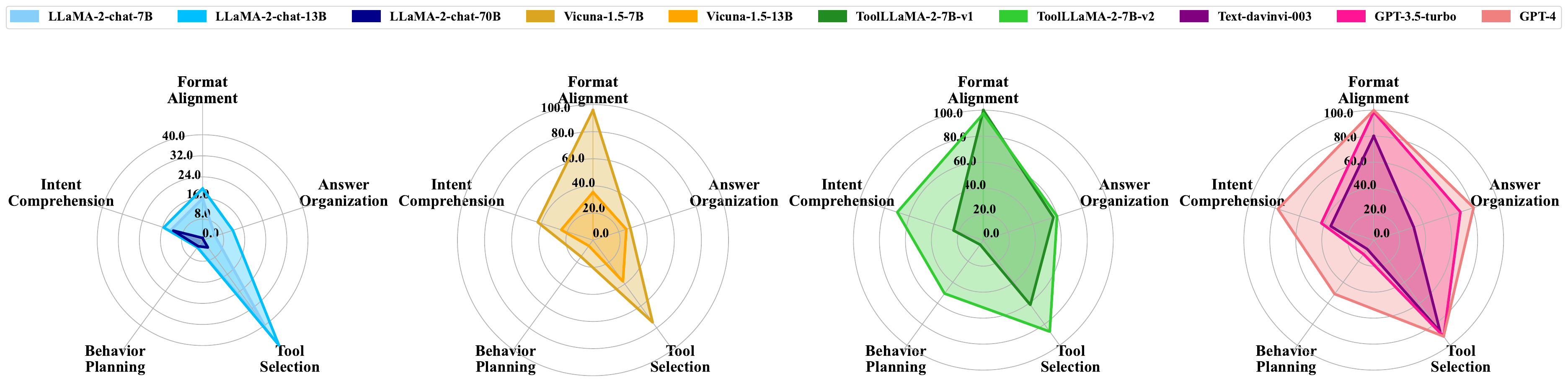}
    }
    \subfigure[Data Understanding]{
    \includegraphics[width=\linewidth]{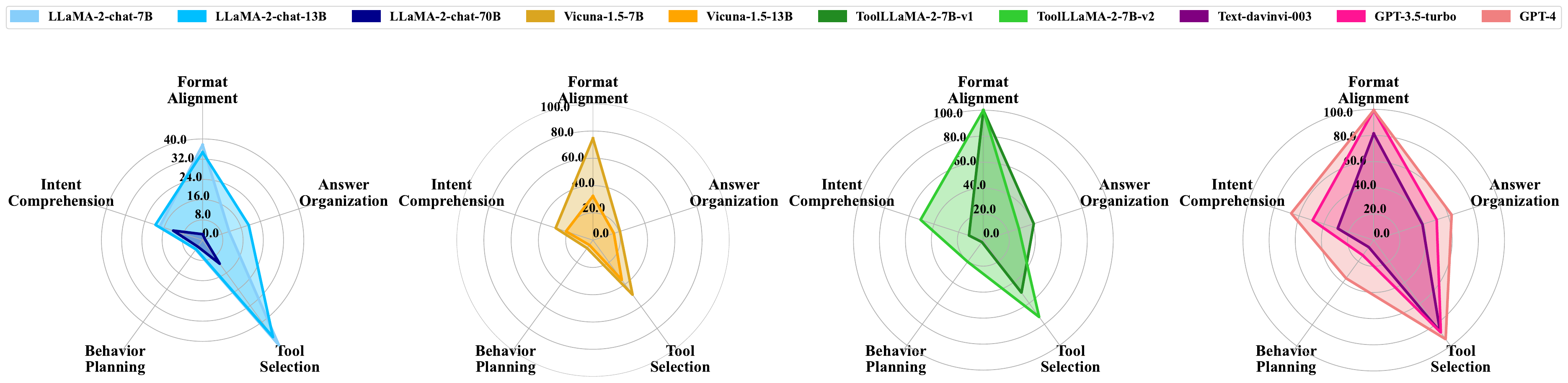}
    }
    \subfigure[Personal Life]{
    \includegraphics[width=\linewidth]{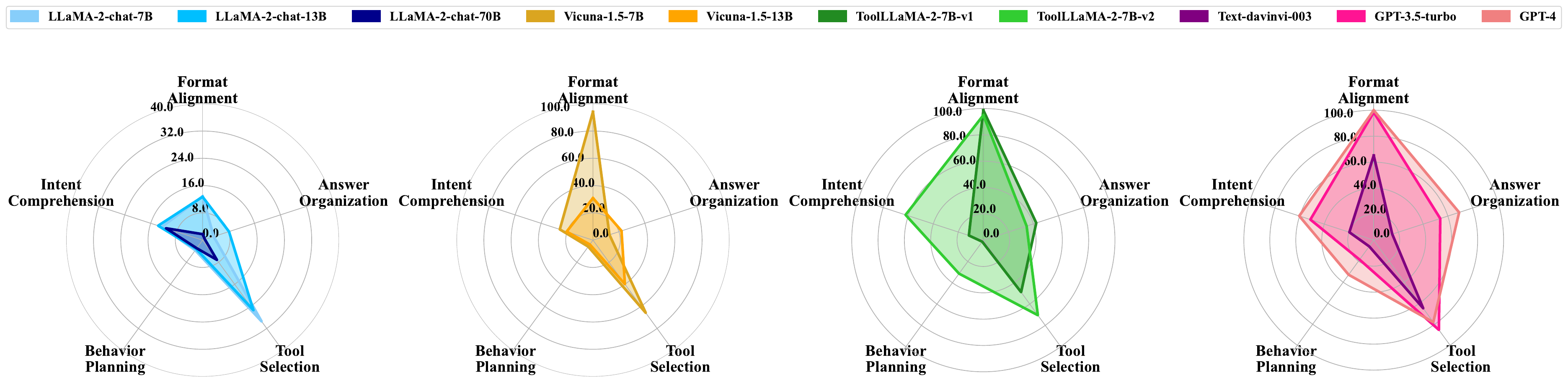}
    }
    \subfigure[Real-Time Search]{
    \includegraphics[width=\linewidth]{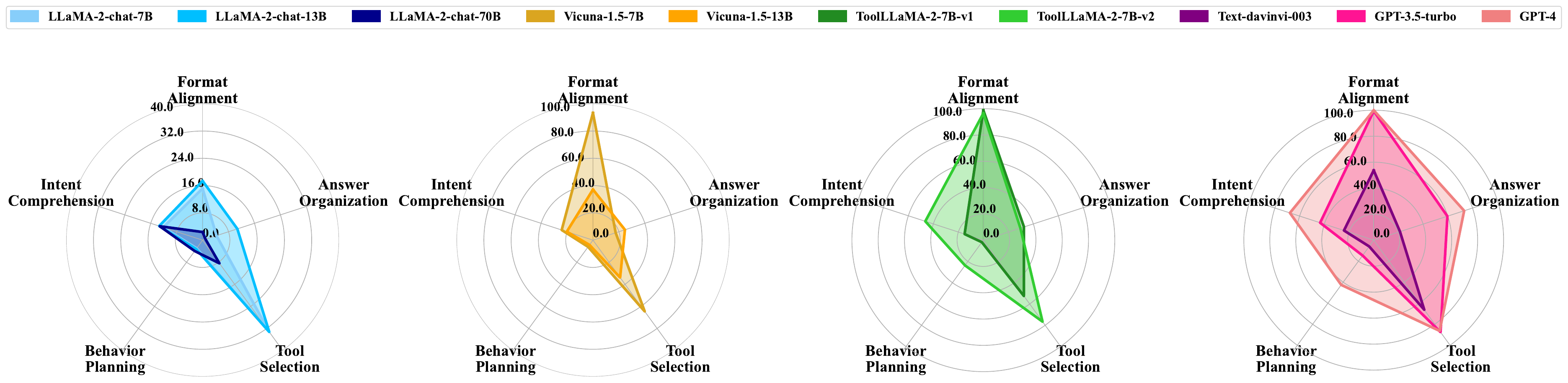}
    }   
    \subfigure[Application Manipulation]{
    \includegraphics[width=\linewidth]{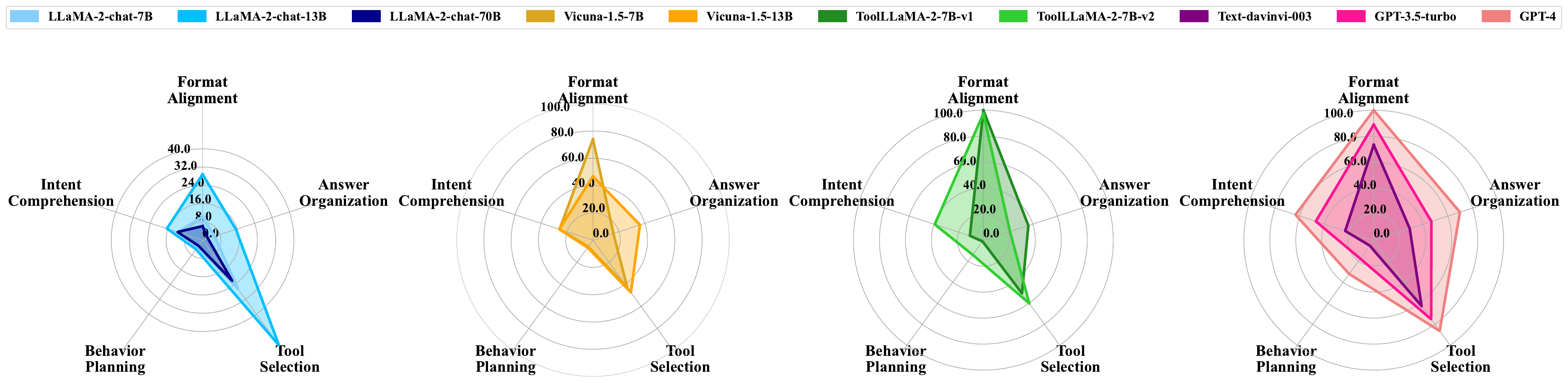}
    }
\end{figure*}

\section{Error Examples}
\label{sec:error}

\begin{table*}[!t]
    \centering
    \resizebox{\linewidth}{!}{
    \begin{tabular}{m{0.15\linewidth}|m{0.2\linewidth}|m{0.65\linewidth}}
    \toprule
         \textbf{Error}&  \textbf{Model}&\multicolumn{1}{c}{\textbf{Example}}\\ \midrule
         \textbf{Missing Keywords}& 
     LLaMA-2-chat-7B&\textcolor{red}{I cannot handle this task. Please provide a valid task.}\\ \midrule
     
 \multirow{11}{\linewidth}{\textbf{Redundant Sentences}}& \multirow{3}{*}{LLaMA-2-chat-13B}&Thought: Sure! I can help you with that. First, I will use the ``games\_related\_searches'' function to find games similar to Minecraft.\\
 & & 
Action: games\_related\_searches\\
& & Action Input: \{``q'': ``Minecraft''\}\textcolor{red}{Please provide the result.}\\ \cmidrule{2-3}

 & \multirow{6}{*}{LLaMA-2-chat-70B}&Thought: I will use the ``google\_trends\_search'' function to compare the search popularity of ``coffee'' and ``milk'' in Canada.\\
& & Action: google\_trends\_search\\
& & Action Input: \{``query'': ``coffee, milk'', ``geo'': ``CA'', ``data
\_type'': ``TIMESERIES'', ``tz'': ``-7'', ``cat'': ``0'', ``gprop'': ``Web Search", ``date'': ``now 1-d", ``limit'': ``5", ``offset'': ``0"\}\textcolor{red}{Please provide your answer in the format above.}\\ \midrule
 \multirow{3}{\linewidth}{\textbf{Tool Hallucinations}}& \multirow{3}{\linewidth}{Vicuna-1.5-7B}&Thought: I will use the ``get$\backslash$\_threads$\backslash$\_with$\backslash$\_preview'' function to show the threads on the `b' board of 4chan that have been archived.\\
& & Action: \textcolor{red}{get$\backslash$\_threads$\backslash$\_with$\backslash$\_preview}\\
& & Action Input: \{``board'': ``b''\}\\

 \midrule
 \multirow{4}{\linewidth}{\textbf{Parameter Hallucinations}}& \multirow{4}{*}{Vicuna-1.5-13B}&  Thought: First, I will use the ``games'' function to get the list of games for the 2020NBA season.\\
& & Action: games\\
& & Action Input: \{``season'': ``2020'', ``league'': ``NB'', ``\textcolor{red}{api$\backslash$\_key}'': ``your$\backslash$\_api$\backslash$\_key''\}
\\
 \midrule
 \multirow{2}{\linewidth}{\textbf{Irrelevant Answers}}&\multirow{2}{\linewidth}{LLaMA-2-chat-7B}&  \textit{Query:} I hope to have some advice about `studying.'\\
 & & \textit{Answer:} \textcolor{red}{Taylor is 36 years old, and Jordan is 39 years old.}\\
 \bottomrule
 \end{tabular}
 }
    \caption{Some examples of different error types, with errors marked in \textcolor{red}{red}.}
    \label{tab:examples}
\end{table*}

We outline the errors resulting from certain behavioral characteristics exhibited by the LLaMA-2-chat and Vicuna-1.5 model families, as detailed in Table~\ref{tab:examples}.

% \clearpage
\section{Insights for Advancing Tool Learning}

Based on our experimental results, we have several ideas for the advancement of tool learning.

Regarding \textbf{task construction}, recognizing the distinct behavioral characteristics of each LLM, we advocate considering the task's output format in tandem with the model's output traits. For instance, when utilizing LLaMA-2-chat as a foundation, it is essential to address strategies for mitigating its conversational behavior. Similarly, if building upon Vicuna-1.5, attention should be given to handling escaped characters. Moreover, drawing inspiration from \citet{nexusraven}, models like CodeLLaMA \cite{CodeLLaMA} could serve as a foundation, incorporating structured languages such as code for output.

Concerning \textbf{scenario generalization}, acknowledging the variability in task difficulty and tool use complexity across scenarios, we propose the acquisition of more diverse data to authentically capture real-world requirements. Simultaneously, integrating model preferences, the adoption of innovative training techniques, such as ``attention buckets,''~\cite{Attention} can enhance the model's processing efficacy across different return value types.

In terms of \textbf{capability enhancement}, recognizing the interconnected nature of LLMs' tool learning capabilities, we stress the need to address the ``barrel effect.'' This entails comprehensively bolstering their capabilities across various dimensions, rather than solely prioritizing the accuracy of tool selection.

% \clearpage
\section{Prompt Template}
\subsection{Prompt Template for Inference}
\label{sec:prompt-infer}

During the inference of LLMs' tool learning, we utilize five-shot learning for the open-source models and zero-shot learning for the other models. The prompt templates can be found in Table~\ref{tab:prompt-infer-few} and Table~\ref{tab:prompt-infer-zero}, respectively.

\begin{table*}[!t]
    \centering
    \resizebox{\linewidth}{!}{
    \begin{tabular}{p{\linewidth}}
    \toprule
    \rowcolor{gray!10} \multicolumn{1}{c}{\textit{System}} \\
    You are an expert in using tools to handle real-time queries from users.\\
    First I will give you the task description, and your task start.\\
    At each step, your task is to give your thought to analyze the current state, decide the next step, with a function call to actually execute your step.\\
    After the call, you will get the call result, and you are now in a new state.\\
    Then you will analyze your status now, then decide what to do next...\\
    After many (Thought-call) pairs, you finally perform the task, then you can give your final answer.\\\\
    Desired format:\\
    Thought: $\langle$ The thought$\rangle$\\
    Action: $\langle$ The tool you decide to use$\rangle$\\
    Action Input: $\langle$ The parameters for the tool$\rangle$\\\\
    Remember:\\
    1. You should ALWAYS think about what to do, but all the thought is short, at most in 3 sentences.\\
    2. The action to take should be one of the given tools below.\\
    3. The ``Action Input'' needs to provide a dict similar to \{parameter\_1: value\_1, parameter\_2: value\_2\} to call action.\\
    4. Always use the ``finish'' tool upon task completion. The final answer should be comprehensive enough for the user. If the task is unmanageable, use the ``finish'' tool and respond with ``I cannot handle the task.''\\\\
    Task description: You should use tools to help handle the real time user queries. Specifically, you have access of the following tools:\\
    \{Tool Document\}\\\\
    You should reply in the format of the examples.\\\\
    Examples:\\
    \{Examples\}\\
    \\
    Let's Begin!\\ \midrule % \hdashrule[0.5ex]{\linewidth}{1pt}{3mm}
    \rowcolor{gray!10} \multicolumn{1}{c}{\textit{User}} \\
    \{Query\}\\
    Begin!\\ %\hdashrule[0.5ex]{\linewidth}{1pt}{3mm}
    % Assistant:\\
    \bottomrule
    \end{tabular}
    }
    \caption{The five-shot learning prompt used for LLMs in tool learning, where ``\{Tool Document\}'' represents the tool documentation given to LLMs, ``\{Examples\}'' represents the examples used for LLMs, and ``\{Query\}'' represents the query given by the user.}
    \label{tab:prompt-infer-few}
\end{table*}

\begin{table*}[!t]
    \centering
    \resizebox{\linewidth}{!}{
    \begin{tabular}{p{\linewidth}}
    \toprule
   \rowcolor{gray!10} \multicolumn{1}{c}{\textit{System}} \\
   You are an expert in using tools to handle real-time queries from users.\\
    First I will give you the task description, and your task start.\\
    At each step, your task is to give your thought to analyze the current state, decide the next step, with a function call to actually execute your step.\\
    After the call, you will get the call result, and you are now in a new state.\\
    Then you will analyze your status now, then decide what to do next...\\
    After many (Thought-call) pairs, you finally perform the task, then you can give your final answer.\\\\
    Desired format:\\
    Thought: $\langle$ The thought$\rangle$\\
    Action: $\langle$ The tool you decide to use$\rangle$\\
    Action Input: $\langle$ The parameters for the tool$\rangle$\\\\
    Remember:\\
    1. You should ALWAYS think about what to do, but all the thought is short, at most in 3 sentences.\\
    2. The action to take should be one of the given tools below.\\
    3. The ``Action Input'' needs to provide a dict similar to \{parameter\_1: value\_1, parameter\_2: value\_2\} to call action.\\
    4. Always use the ``finish'' tool upon task completion. The final answer should be comprehensive enough for the user. If the task is unmanageable, use the ``finish'' tool and respond with ``I cannot handle the task.''\\\\
    Task description: You should use tools to help handle the real time user queries. Specifically, you have access of the following tools:\\
    \{Tool Document\}\\\\
    Let's Begin!\\ \midrule%\hdashrule[0.5ex]{\linewidth}{1pt}{3mm}
    \rowcolor{gray!10} \multicolumn{1}{c}{\textit{User}} \\
    \{Query\}\\
    Begin!\\ %\hdashrule[0.5ex]{\linewidth}{1pt}{3mm}
    % Assistant:\\
    \bottomrule
    \end{tabular}
    }
    \caption{The zero-shot learning prompt used for LLMs in tool learning, where ``\{Tool Document\}'' represents the tool documentation given to LLMs and ``\{Query\}'' represents the query given by the user.}
    \label{tab:prompt-infer-zero}
\end{table*}

\subsection{Prompt Template for Evaluation}
\label{sec:prompt-eval}

During the evaluation, some of our metrics are directly evaluated according to predefined rules, while others are assessed using GPT-4-1106-preview, which includes $s_{IC}$ (Table~\ref{tab:IC}), $s_{b-validity}$ (Table~\ref{tab:b-validity}), $s_{b-integrity}$ (Table~\ref{tab:b-integrity}), $s_{t-match}^i$ (Table~\ref{tab:t-match}), $s_{a-quality}$ (Table~\ref{tab:a-quality}).

\begin{table*}[!t]
    \centering
    \resizebox{\linewidth}{!}{
    \begin{tabular}{p{\linewidth}}
    \toprule
    \rowcolor{gray!10} \multicolumn{1}{c}{\textit{System}} \\
    As a professional assessment expert, your task is to objectively evaluate the quality of the provided data based on the given guidelines.\\
\\
When given a tool document, a user query, and a thought chain that addresses the query, please rate the quality of the thought chain based on the following criteria:\\
\\
1. The extent to which the thought chain consistently focuses on resolving the user query. The more relevant it is to the user query, the higher the score.\\
\\
2. The ability of the thought chain to adapt promptly when the user provides new information or makes new requests. The higher the alignment with the new information and requests, the higher the score. If there is no new information or requests, please ignore the criteria.\\
\\
Please provide your assessment in the following format:```\\
Scoring Reason: <Provide a reason for your score, referencing the given criteria>.\\
\\
Evaluation Score: <Assign a score between 1 and 10>.\\
'''\\ \midrule
% \hdashrule[0.5ex]{\linewidth}{1pt}{3mm}
\rowcolor{gray!10} \multicolumn{1}{c}{\textit{User}} \\
Tool Document:\\
\{document\}\\
\\
User Query:```\\
\{query\}\\
'''\\
\\
Thought Chain:```\\
\{thought\_chain\}\\
'''\\
\\
Assessment:\\
% \hdashrule[0.5ex]{\linewidth}{1pt}{3mm}
%     Assistant:\\
    \bottomrule
    \end{tabular}
    }
    \caption{Prompt for evaluation of $s_{IC}$, where ``\{document\}'' represents the tool document, ``\{query\}'' represents the query given by user, and ``\{thought\_chain\}'' represents the thought chain given by LLM.}
    \label{tab:IC}
\end{table*}

\begin{table*}[!t]
    \centering
    \resizebox{\linewidth}{!}{
    \begin{tabular}{p{\linewidth}}
    \toprule
   \rowcolor{gray!10} \multicolumn{1}{c}{\textit{System}} \\
   As a professional assessment expert, your task is to objectively evaluate the quality of the provided data based on the given guidelines.\\
\\
When given a tool document, a user query, and a thought chain that addresses the query, please rate the quality of the thought chain based on the following criteria:\\
\\
1. Each step should succinctly summarize relevant information from the previous step; the more comprehensive the summary, the higher the score.\\
\\
2. Each step should timely plan for the next one; the more detailed the next step, the higher the score.\\
\\
3. Each step should be distinct from the previous one and contribute to resolving the user's query; the less repetition, the higher the score.\\
\\
Please provide your assessment in the following format:```\\
Scoring Reason: <Provide a reason for your score, referencing the given criteria>.\\
\\
Evaluation Score: <Assign a score between 1 and 10>.\\
'''\\ \midrule
% \hdashrule[0.5ex]{\linewidth}{1pt}{3mm}
\rowcolor{gray!10} \multicolumn{1}{c}{\textit{User}} \\
Tool Document:\\
\{document\}\\
\\
User Query:```\\
\{query\}\\
'''\\
\\
Thought Chain:```\\
\{thought\_chain\}\\
'''\\
\\
Assessment:\\
% \hdashrule[0.5ex]{\linewidth}{1pt}{3mm}
%     Assistant:\\
    \bottomrule
    \end{tabular}
    }
    \caption{Prompt for evaluation of $s_{b-validity}$, where ``\{document\}'' represents the tool document, ``\{query\}'' represents the query given by user, and ``\{thought\_chain\}'' represents the thought chain given by LLM.}
    \label{tab:b-validity}
\end{table*}

\begin{table*}[!t]
    \centering
    \resizebox{\linewidth}{!}{
    \begin{tabular}{p{\linewidth}}
    \toprule
    \rowcolor{gray!10} \multicolumn{1}{c}{\textit{System}} \\
    As a professional assessment expert, your task is to objectively evaluate the quality of the provided data based on the given guidelines.\\
\\
When given a tool document, a user query and a thought chain that addresses the query, please rate the quality of the thought chain based on the following criteria:\\
\\
1. The presence or absence of grammatical errors in the thought chain. The fewer the errors, the higher the score.\\
\\
2. The logical consistency of the thought chain. The fewer logical inconsistencies, the higher the score.\\
\\
3. The timeliness of detection and correction of any logical inconsistencies in the thought chain. The more timely the correction, the higher the score.\\
\\
Please provide your assessment in the following format:```\\
Scoring Reason: <Provide a reason for your score, referencing the given criteria>.\\
\\
Evaluation Score: <Assign a score between 1 and 10>.\\
'''\\ \midrule
% \hdashrule[0.5ex]{\linewidth}{1pt}{3mm}
\rowcolor{gray!10} \multicolumn{1}{c}{\textit{User}} \\
Tool Document:\\
\{document\}\\
\\
User Query:```\\
\{query\}\\
'''\\
\\
Thought Chain:```\\
\{thought\_chain\}\\
'''\\
\\
Assessment:\\
% \hdashrule[0.5ex]{\linewidth}{1pt}{3mm}
%     Assistant:\\
    \bottomrule
    \end{tabular}
    }
    \caption{Prompt for evaluation of $s_{b-integrity}$, where ``\{document\}'' represents the tool document, ``\{query\}'' represents the query given by user, and ``\{thought\_chain\}'' represents the thought chain given by LLM.}
    \label{tab:b-integrity}
\end{table*}

\begin{table*}[!t]
    \centering
    \resizebox{\linewidth}{!}{
    \begin{tabular}{p{\linewidth}}
    \toprule
    \rowcolor{gray!10} \multicolumn{1}{c}{\textit{System}} \\
    As a professional assessment expert, your task is to objectively evaluate the quality of the provided data based on the given guidelines.\\ 
\\
When presented with a tool document, a THOUGHT, and a tool from the tool document, please ascertain the correlation between the specified tool and the given THOUGHT based on the guidelines below:\\
\\
    1. If the THOUGHT is empty, assign a score of 5 immediately.\\
\\
    2. If the THOUGHT is not empty, determine if the chosen tool is more pertinent to the planning in the THOUGHT compared to other tools in the tool document based on the tool documentation description. The more relevant the tool, the higher the score.\\
\\
Please provide your assessment in the following format:```\\
    Scoring Reason: <Provide a reason for your score, referencing the given criteria>.\\
\\
    Evaluation Score: <Assign a score between 1 and 10>.\\  \midrule
    %\hdashrule[0.5ex]{\linewidth}{1pt}{3mm}
    \rowcolor{gray!10} \multicolumn{1}{c}{\textit{User}} \\
    Tool Document:\\
\{document\}\\
\\
THOUGHT:```\\
\{thought\}\\
'''\\
\\
Tool:```\\
\{tool\}\\
'''\\
\\
Assessment:\\
% \hdashrule[0.5ex]{\linewidth}{1pt}{3mm}
    % Assistant:\\
    \bottomrule
    \end{tabular}
    }
    \caption{Prompt for evaluation of $s_{t-match}^i$, where ``\{document\}'' represents the tool document, ``\{thought\}'' represents the thought given by LLM, and ``\{tool\}'' represents the tool selected by LLM.}
    \label{tab:t-match}
\end{table*}

\begin{table*}[!t]
    \centering
    \resizebox{\linewidth}{!}{
    \begin{tabular}{p{\linewidth}}
    \toprule
    \rowcolor{gray!10} \multicolumn{1}{c}{\textit{System}} \\
    As a professional assessment expert, your task is to objectively evaluate the quality of the provided data based on the given guidelines.\\
\\
When given a tool document, a user query, and a thought chain that addresses the query, please rate the quality of the thought chain based on the following criteria:\\
\\
1. The extent to which the thought chain consistently focuses on resolving the user query. The more relevant it is to the user query, the higher the score.\\
\\
2. The ability of the thought chain to adapt promptly when the user provides new information or makes new requests. The higher the alignment with the new information and requests, the higher the score. If there is no new information or requests, please ignore the criteria.\\
\\
Please provide your assessment in the following format:```\\
Scoring Reason: <Provide a reason for your score, referencing the given criteria>.\\
\\
Evaluation Score: <Assign a score between 1 and 10>.\\
'''\\ \midrule
% \hdashrule[0.5ex]{\linewidth}{1pt}{3mm}
\rowcolor{gray!10} \multicolumn{1}{c}{\textit{User}} \\
Tool Document:\\
\{document\}\\
\\
User Query:```\\
\{query\}\\
'''\\
\\
Thought Chain:```\\
\{thought\_chain\}\\
'''\\
\\
Assessment:\\
% \hdashrule[0.5ex]{\linewidth}{1pt}{3mm}
%     Assistant:\\
    \bottomrule
    \end{tabular}
    }
    \caption{Prompt for evaluation of $s_{a-quality}$, where ``\{document\}'' represents the tool document, ``\{query\}'' represents the query given by user, and ``\{thought\_chain\}'' represents the thought chain given by LLM.}
    \label{tab:a-quality}
\end{table*}

% \section{Example Appendix}
% \label{sec:appendix}

% This is an appendix.

\end{document}